\documentclass[10pt,twocolumn,letterpaper]{article}

\usepackage[pagenumbers]{cvpr}      

\usepackage[utf8]{inputenc} 
\usepackage[T1]{fontenc}    
\usepackage{microtype}      

\definecolor{cvprblue}{rgb}{0.21,0.49,0.74}
\PassOptionsToPackage{pagebackref,breaklinks,colorlinks,allcolors=cvprblue}{hyperref}
\usepackage{doi}            
\usepackage{xurl}           
\usepackage{makecell} 
\usepackage{pifont} 
\usepackage{multirow}

\def\paperID{*****} 
\def\confName{CVPR}
\def\confYear{2026}

\definecolor{okgreen}{HTML}{009B3A}
\definecolor{partgold}{HTML}{D99A00}
\definecolor{badred}{HTML}{E00000}
\definecolor{headerink}{HTML}{000000}
\definecolor{accentblue}{HTML}{0050A4}
\newcommand{\Y}{\textcolor{okgreen}{\ding{51}}}
\newcommand{\Ym}{\textcolor{partgold}{\ding{51}}}
\newcommand{\N}{\textcolor{badred}{\ding{55}}}
\newcommand{\hd}[1]{\textcolor{headerink}{\bfseries\fontsize{7.5}{8.5}\selectfont #1}}

\title{PhysCodeGym: Benchmarking Coding Agents for Physically Verifiable 4D Reconstruction}
\makeatletter
\let\appendixpapertitle\@title
\makeatother

\author{%
  Shuai Shao\textsuperscript{1,}\thanks{Equal contribution. All four co-first authors agree that each may list their own name first when presenting this work.}\textsuperscript{,}\thanks{Project lead. \href{mailto:shaoshuai.ederson@sjtu.edu.cn}{\nolinkurl{shaoshuai.ederson@sjtu.edu.cn}}.}, Jinchen He\textsuperscript{1,*}, Cunxin Fan\textsuperscript{1,*}, Zihan Guo\textsuperscript{2,3,*,}\thanks{Co-corresponding authors. Weinan Zhang: \href{mailto:wnzhang@sjtu.edu.cn}{\nolinkurl{wnzhang@sjtu.edu.cn}}.},\\
  Shijian Wang\textsuperscript{4}, Qihan Ren\textsuperscript{1}, Kangning Zhang\textsuperscript{1}, Zunnan Xu\textsuperscript{5}, Haoyuan Li\textsuperscript{2},\\
  Hao Wang\textsuperscript{6}, Jianghao Lin\textsuperscript{1,\textdaggerdbl}, Weiwen Liu\textsuperscript{1,\textdaggerdbl}, Weinan Zhang\textsuperscript{1,\textdaggerdbl}\\[4pt]
  {\normalsize \textsuperscript{1}Shanghai Jiao Tong University\quad \textsuperscript{2}Sun Yat-sen University\quad \textsuperscript{3}Shanghai Innovation Institute}\\
  {\normalsize \textsuperscript{4}Southeast University\quad \textsuperscript{5}Nanyang Technological University\quad \textsuperscript{6}Carnegie Mellon University}
}

\hypersetup{
pdftitle={PhysCodeGym: Benchmarking Coding Agents for Physically Verifiable 4D Reconstruction},
pdfauthor={Shuai Shao, Jinchen He, Cunxin Fan, Zihan Guo, Shijian Wang, Qihan Ren, Kangning Zhang, Zunnan Xu, Haoyuan Li, Hao Wang, Jianghao Lin, Weiwen Liu, Weinan Zhang},
}

\makeatletter
\apptocmd{\@maketitle}{
\noindent\begin{minipage}{\textwidth}
  \centering
  \makebox[\textwidth][c]{%
    \includegraphics[width=1.074\textwidth,pagebox=cropbox]{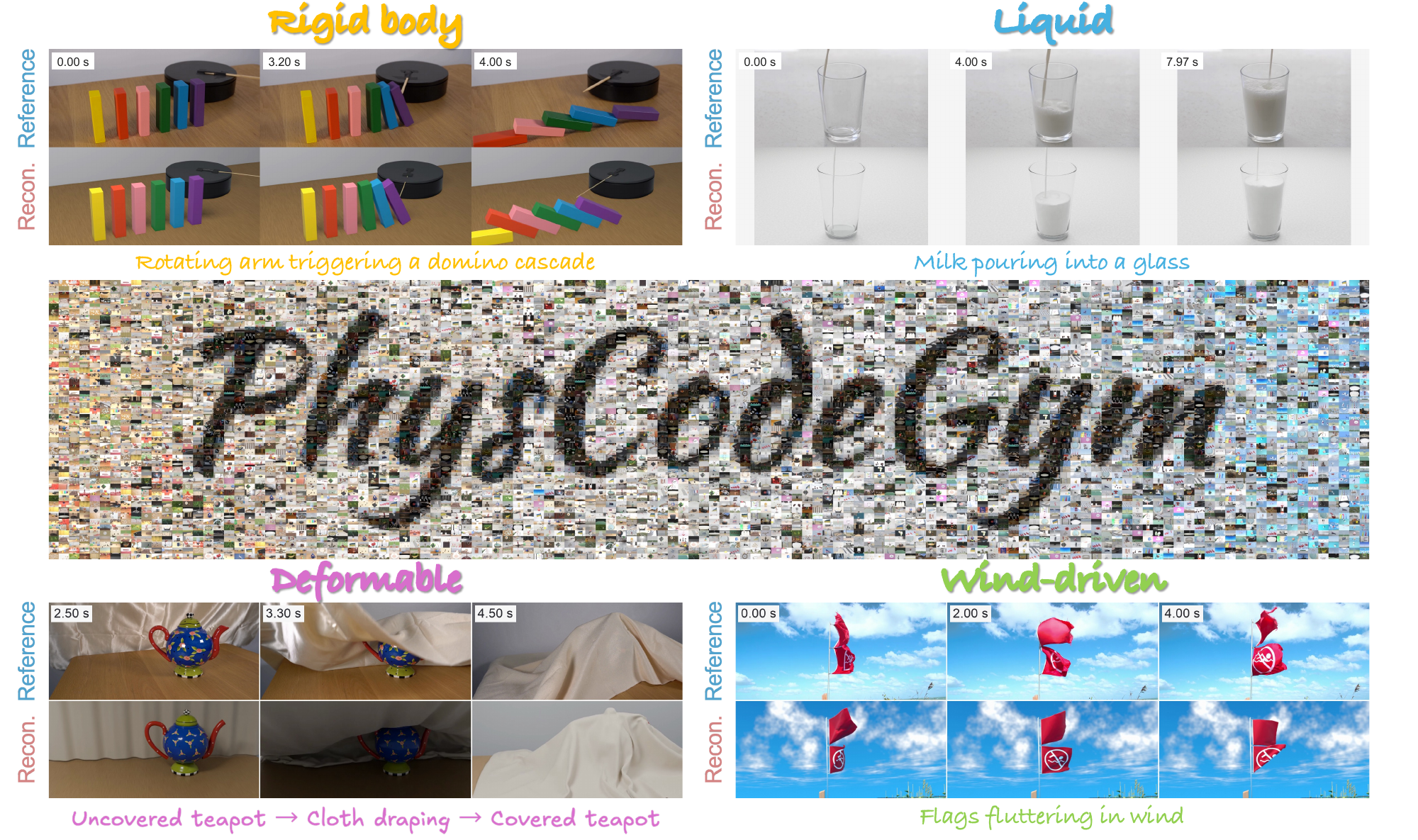}%
  }
  \captionsetup{hypcap=false}
  \captionof{figure}{\textbf{PhysCodeGym} benchmarks coding agents on $156$ video-to-Blender reconstruction tasks
  spanning rigid-body, liquid, deformable, and wind-driven motion.
  Examples pair reference frames (top) with agent reconstructions (bottom).
  Evaluation measures visual fidelity, dynamic behavior, and physical validity.}
  \label{fig:teaser}
\end{minipage}
\par\vspace{3pt}
}{}{}
\makeatother

\makeatletter
\pretocmd{\@maketitle}{%
  \def\@makefnmark{\hbox{\@textsuperscript{\normalfont\@thefnmark}}}%
}{}{}
\makeatother

\begin{document}
\raggedbottom
\addtocontents{toc}{\protect\setcounter{tocdepth}{-1}}

\maketitle

\begin{abstract}

Building physical simulations from video requires coding agents to translate observations into executable models and refine them through tool use.
We introduce PhysCodeGym, a benchmark for physically verifiable 4D reconstruction with $156$ tasks covering rigid-body, liquid, deformable, and wind-driven motion.
Agents autonomously inspect videos, write and run Blender programs, and revise their scenes using simulation feedback.
Six metrics combine rule-based measurements with rubric-based judgments to assess visual and physical reconstruction quality.
Program inspection, scene measurements, and counterfactual interventions test physical consistency under original and modified conditions.
Experiments with frontier coding agents reveal a tradeoff between artifact quality and completion across reasoning-effort settings, while harness comparisons and agent traces link verification behavior with reconstruction performance.
Case studies show that matching visible objects and events can coexist with physical defects.
Counterfactual tests further expose latent contact and support failures in reconstructions that pass ordinary physical checks, highlighting the gap between reproducing observations and building reliable physical models.
Code is available at \url{https://github.com/ShaoShuai0605/PhysCodeGym}.

\end{abstract}


\section{Introduction}

Physics-aware reconstruction enables the creation of simulation-ready objects and scenes for applications in animation~\citep{DBLP:conf/cvpr/XieZQLF0J24}, gaming~\citep{DBLP:conf/siggraph/JiangYXLFWLLG0J24}, manufacturing~\citep{DBLP:conf/iclr/XiongHL0G025}, and robot learning~\citep{lou2026DREX}.
Specialized methods estimate scene geometry and physical properties within predefined representations and simulation pipelines~\citep{DBLP:conf/iclr/LiQCJLJG23,DBLP:conf/cvpr/ChenDWHCFGL25,DBLP:conf/nips/WuYLFT15,DBLP:conf/nips/0001LKFT17}.
More recently, coding agents have emerged as a promising path forward by generating executable scene programs and revising them through execution and visual feedback~\citep{DBLP:conf/eccv/YinGWWLBDKF26,wangCodeWorldsAgentic2026}.

Recent frontier models, such as GPT-6 Astra, demonstrate substantially improved capabilities in executable 4D scene construction, creating an urgent need to quantitatively assess their physical fidelity.
We argue that a benchmark for this purpose should meet three requirements.
First, \textbf{broad interaction coverage} should span diverse processes such as contact, fluid flow, and deformation~\citep{DBLP:conf/iclr/BansalLXZYBJSCG25}.
Second, \textbf{hybrid assessments} should combine rule-based checks with rubric-based judgments~\citep{yaoApple$p$BenchmarkingThinking2026}.
Rule-based checks provide precise measurements but are limited to predefined criteria.
Rubric-based judgments can assess broader task requirements~\citep{DBLP:conf/acl/LeeKPKS24,DBLP:conf/eccv/PothirajCZSB26}, though they may miss physical violations~\citep{liuSimWorldsMultiagentSystem2026,motamedTRAVLRecipeMaking2025}.
Third, agreement with observed motion and appearance does not ensure correct behavior under altered physical conditions~\citep{DBLP:conf/iclr/BaradelNMM020,DBLP:conf/iclr/YiGLK0TT20,shen4DCodeBenchBenchmarkingAgents2026,wangCodeWorldsAgentic2026,DBLP:conf/iclr/ZhangJDLUZWWPLK26,caiWhatifWorldCausal2026}, motivating \textbf{factual and counterfactual evaluation}. For example, hardcoded motion trajectories may make balls appear to collide even when contact is disabled.
While prior benchmarks cover subsets of these design choices, we argue that considering them jointly provides a holistic view of the capabilities and limitations of frontier coding agents in 4D reconstruction.

To this end, we introduce PhysCodeGym, a benchmark for evaluating coding agents on physically verifiable 4D reconstruction (\cref{fig:teaser}).
Given a reference video, an agent writes a standalone Blender script to create an editable scene and render its animation.
Our benchmark comprises 156 tasks spanning rigid-body, liquid, deformable, and wind-driven dynamics, with scenarios involving collisions, suspension, pouring, and cloth deformation.

To enable systematic evaluation across all 156 tasks (\cref{sec:evaluation}), we provide object checklists, 578 timestamped event annotations, and task-specific physical requirements.
The evaluation protocol combine these annotations with LLM-based and rule-based judges to compute six metrics.
Five assess the submitted video for structural and perceptual similarity, object accuracy, event completion, and motion overlap.
To support event completion scoring, we specify failure and uncertainty criteria and aggregate three independent judgments per submission by majority vote.
The sixth metric, physical validity, is assessed through a task-specific measurement that combines code and scene inspection with frame-by-frame analysis of native states, contacts, and forces.
The scoring further incorporates counterfactual checks, in which copies of the scene are rerun under altered physical conditions.
For example, our protocol redirects a billiard ball toward the table edge to check whether it falls after losing support.
To verify that our protocol reliably detects physical violations, we perform manual checks by perturbing submitted scenes and injecting known defects.

We evaluate nine coding agents to characterize their strengths and failure modes in 4D physical reconstruction. We find that, while these agents can reconstruct recognizable objects, they often struggle to accurately capture motion and physical interactions (\cref{tab:results}).
Moreover, reconstructions with perfect object and event scores can violate physical constraints, including energy conservation (\cref{sec:results-analysis}).
Our harness-level comparison further reveals that, at high effort, Codex achieves higher visual and event scores, while Claude Code has slightly higher motion overlap (\cref{fig:cross-harness}).
Reasoning effort presents an additional tradeoff: low effort completes more tasks and leads overall under a fixed budget, whereas max effort scores higher among successful tasks (\cref{sec:results-analysis}).

Our contributions are:
{\begin{itemize}[leftmargin=*, nosep]
\item \textbf{A benchmark for coding agents that building physical simulations.}
PhysCodeGym uses $156$ tasks to test coding agents’ ability to build physics-aware Blender scenes.
The tasks cover gravity and support, contact and friction, fluids, deformation, and wind-driven motion.
\item \textbf{A protocol for joint visual and physical evaluation.}
We combine LLM-based and rule-based judges with factual and counterfactual checks.
Our diagnostic cases include reconstructions that pass object and event checks but fail native-scene audits or targeted interventions.
\item \textbf{An empirical study of coding agents for physical reconstruction.}
We evaluate leading agents on PhysCodeGym and study how their results vary with harness choice and configured reasoning effort. We highlight their strengths, failure modes and shortcut behaviors, informing future development of physically aware coding agents.
\end{itemize}}


\begin{table*}[!t]
\centering
\caption{Comparison of programmatic scene generation and reconstruction benchmarks.
\textbf{Task} distinguishes generation (Gen.) from reconstruction (Recon.); \textbf{Input} specifies the conditioning modality and source (Syn.: synthetic).
\textbf{Visual Eval.} assesses rendered appearance or spatial layout.
Under \textbf{Behavioral Fidelity}, \textbf{Event Eval.} checks target actions or outcomes, and \textbf{Motion Alignment} compares motion against a reference.
Under \textbf{Physical Implementation}, \textbf{Internal Physics Checks} inspect scene structures, physical parameters, or simulation states, and \textbf{Mechanism Dependence} uses matched physical interventions to test whether the modeled mechanisms drive the observed behavior.
\textbf{Harness} lists reported execution workflows or interfaces.
\Y\ included, \Ym\ limited scope, \N\ not evaluated under the column definition.}
\label{tab:comparison}
\fontsize{8}{9}\selectfont
\renewcommand{\hd}[1]{\textcolor{headerink}{\bfseries\fontsize{8}{9}\selectfont #1}}
\renewcommand{\arraystretch}{1.10}
\setlength{\tabcolsep}{1pt}
\setlength{\aboverulesep}{1.5pt}
\setlength{\belowrulesep}{1.5pt}
\setlength{\heavyrulewidth}{0.8pt}
\setlength{\lightrulewidth}{0.4pt}
\begin{tabular*}{\textwidth}{@{\extracolsep{\fill}}l c c c c c c c l@{}}
\toprule
\multirow{2}{*}{\hd{Benchmark}} &
\multirow{2}{*}{\hd{Task}} &
\multirow{2}{*}{\hd{Input}} &
\multirow{2}{*}{\hd{\makecell{Visual\\Eval.}}} &
\multicolumn{2}{c}{\hd{Behavioral Fidelity}} &
\multicolumn{2}{c}{\hd{Physical Implementation}} &
\multicolumn{1}{c}{\multirow{2}{*}{\hd{Harness}}} \\
\cmidrule(lr){5-6}\cmidrule(lr){7-8}
& & & & \hd{\makecell{Event\\Eval.}} &
\hd{\makecell{Motion\\Alignment}} &
\hd{\makecell{Internal\\Physics Checks}} &
\hd{\makecell{Mechanism\\Dependence}} & \\
\midrule
Code4D~\citep{zhangCode2WorldsEmpoweringCoding2026a} & Gen. & Text & \Y & \N & \N & \N & \N & Multi-agent code loop \\
4DBuildBench~\citep{liuSimWorldsMultiagentSystem2026} & Gen. & Text & \Y & \Y & \N & \Y & \N & Multi-agent MCP \\
PhysCodeBench~\citep{xiePhysCodeBenchBenchmarkingPhysicsaware2026} & Gen. & Text & \Y & \Ym & \N & \Y & \N & Direct codegen. / SMRF \\
SceneActBench~\citep{zhaoSceneActBenchCanAgents2026} & Recon. & Syn. frames & \Ym & \N & \Y & \N & \N & Shared MCP \\
VisPhyBench~\citep{liangVisPhyWorldProbingPhysical2026} & Recon. & Syn. frames & \Y & \N & \Y & \Ym & \N & Codegen. + repair \\
MPMWorlds~\citep{kovacicMPMWorldsMaterialpointmethodSimulations2026} & Recon. & Syn. video & \Y & \N & \Y & \N & \N & Scaffolded codegen. \\
BVB~\citep{tangBVBBenchmarkingAgentic2026} & Recon. & Real video & \Y & \N & \Ym & \N & \N & Mini-BVB \\
4DCodeBench~\citep{shen4DCodeBenchBenchmarkingAgents2026} & Recon. & \makecell{Real + Syn.\\video} & \Y & \Y & \Y & \Ym & \N & Native CLIs / Stirrup \\
\midrule
\rule{0pt}{9pt}\textbf{Ours} & \textbf{Recon.} & \textbf{Video} & \Y & \Y & \Y & \Y & \Y & \textcolor{accentblue}{\textbf{Native CLIs (all models)}} \\
\bottomrule
\end{tabular*}
\end{table*}

\section{Related Work}

\paragraph{Video-based Physical Reconstruction.}
Video-based physical reconstruction requires recovering scene geometry, physical properties, and dynamics from observations~\citep{rhoMonoPhysicsEstimatingGeometry2026,yue2026videophysedit}.
Existing methods couple geometry and appearance representations with differentiable continuum simulation, mesh-free reduced-order simulation, or spring--mass dynamics to reconstruct simulatable scenes for subsequent prediction and re-simulation~\citep{DBLP:conf/iclr/LiQCJLJG23,DBLP:conf/cvpr/ChenDWHCFGL25,DBLP:conf/iccv/JiangHZYWL25,DBLP:conf/eccv/ZhongYWL24}.
Language models also support physical scene construction: $\Delta$YNAMICS generates structured scene configurations, evaluates reconstruction through MuJoCo re-simulation, and supports editing gravity and initial velocities~\citep{DBLP:conf/cvpr/KaoHWVSHOZ26}.
OVOW recovers instance meshes, poses, and deformations from monocular video, using ground-plane estimation and contact projection to assemble scenes for downstream simulation~\citep{DBLP:conf/eccv/ChenZCZZZZHLW26}.
These methods study how to recover simulatable scenes from video.
PhysCodeGym turns this reconstruction objective into an evaluation task for coding agents, testing whether their generated scene programs satisfy both reference observations and task-specific physical requirements.

\paragraph{Programmatic Scene Construction and Evaluation.}
Earlier inverse-graphics systems represented scenes with probabilistic programs in Picture and structured scene descriptions in Neural Scene De-rendering~\citep{DBLP:conf/cvpr/KulkarniKTM15,DBLP:conf/cvpr/0001TK17}.
Programmatic methods perform 3D generation and editing through code~\citep{DBLP:conf/3dim/SunHDWQG25,DBLP:conf/icml/HuIJKYRSF24,DBLP:conf/eccv/HuangYG24,DBLP:conf/cvpr/ZhangLZW025,DBLP:conf/naacl/YamadaCLHY025,duBlenderLLMTrainingLarge2024,DBLP:conf/wacv/WangZTWDLCBHWR26,kulits2024rethinking}, with benchmarks covering graphics editing, text- or image-conditioned 3D modeling, and image-to-scene inverse rendering~\citep{DBLP:conf/cvpr/GuHJYG25,gao3DCodeBenchBenchmarkingAgentic2026,DBLP:conf/nips/LiuLLWLYZLHF25,dorisCADBenchMultimodalBenchmark2026,zhengVoxelCodeBenchBenchmarking3D2026}.
Thinking in Blender reconstructs single images as editable Blender programs by progressively refining geometry, materials, composition, and lighting~\citep{heThinkingBlenderStaged2026}.
VIGA extends code iteration and visual feedback to dynamic scenes~\citep{DBLP:conf/eccv/YinGWWLBDKF26}, while Code2Worlds generates dynamic scenes from text and refines them through motion critique and code revision~\citep{zhangCode2WorldsEmpoweringCoding2026a}.
For physical implementation, SimWorlds' 4DBuildBench separately evaluates visual output and internal physics, auditing solver configurations, constraints, and animation substitutes~\citep{liuSimWorldsMultiagentSystem2026}; PhysCodeBench evaluates simulation behavior through physical assertions and conservation diagnostics read from engine states~\citep{xiePhysCodeBenchBenchmarkingPhysicsaware2026}.
These dynamic generation benchmarks check scenes and physical implementations against textual requirements; video reconstruction additionally requires agreement with reference observations of appearance and behavior.

\paragraph{Video-to-Program Reconstruction Benchmarks.}
Video-to-program benchmarks evaluate agents' ability to recover scene structure and dynamic behavior.
BVB reconstructs real videos as Blender animations, evaluating spatiotemporal fact retention and representation-level similarity of layout and dynamics under a protocol that prohibits physics simulation~\citep{tangBVBBenchmarkingAgentic2026}.
SceneActBench's dynamic task evaluates layout and keyframed trajectory recovery with supplied assets~\citep{zhaoSceneActBenchCanAgents2026}.
VisPhyWorld and MPMWorlds recover executable simulations from synthetic visual observations, studying physical-parameter and motion reconstruction, and future prediction of multi-material dynamics, respectively~\citep{liangVisPhyWorldProbingPhysical2026,kovacicMPMWorldsMaterialpointmethodSimulations2026}.
The closely related 4DCodeBench reconstructs real and synthetic videos as executable dynamic scenes, evaluating appearance, geometry, and motion alongside state/event questions and mesh-integrity diagnostics, with direct 3D ground-truth comparisons on its synthetic subset.
However, its task permits keyframes, prescribed motion, and physical simulation as alternative dynamic representations, primarily evaluating submissions by reconstruction fidelity~\citep{shen4DCodeBenchBenchmarkingAgents2026}.
Code as Worlds constructs, verifies, and refines executable worlds from text or real videos, supports re-execution under changed physical conditions, and uses these worlds for physical-reasoning supervision~\citep{wangCodeWorldsAgentic2026}.
Its emphasis is on discovering executable worlds and using them for physical reasoning; PhysCodeGym instead makes the physical implementation of core motion a requirement of video reconstruction and tests mechanism dependence through independent evaluation of submitted artifacts.
Evaluation combines reference-event and motion alignment, scene and runtime checks, and matched interventions on contacts, support, or actuation, comparing behavior before and after intervention to determine whether reconstructed motion depends on task-required physical mechanisms.

Table~\ref{tab:comparison} compares representative benchmarks by task, input, visual evaluation, behavioral fidelity, physical implementation evaluation, and execution harness.

\section{PhysCodeGym: Dataset and Task}
\label{sec:benchmark}

\begin{figure*}[!t]
  \centering
  \includegraphics[width=\textwidth]{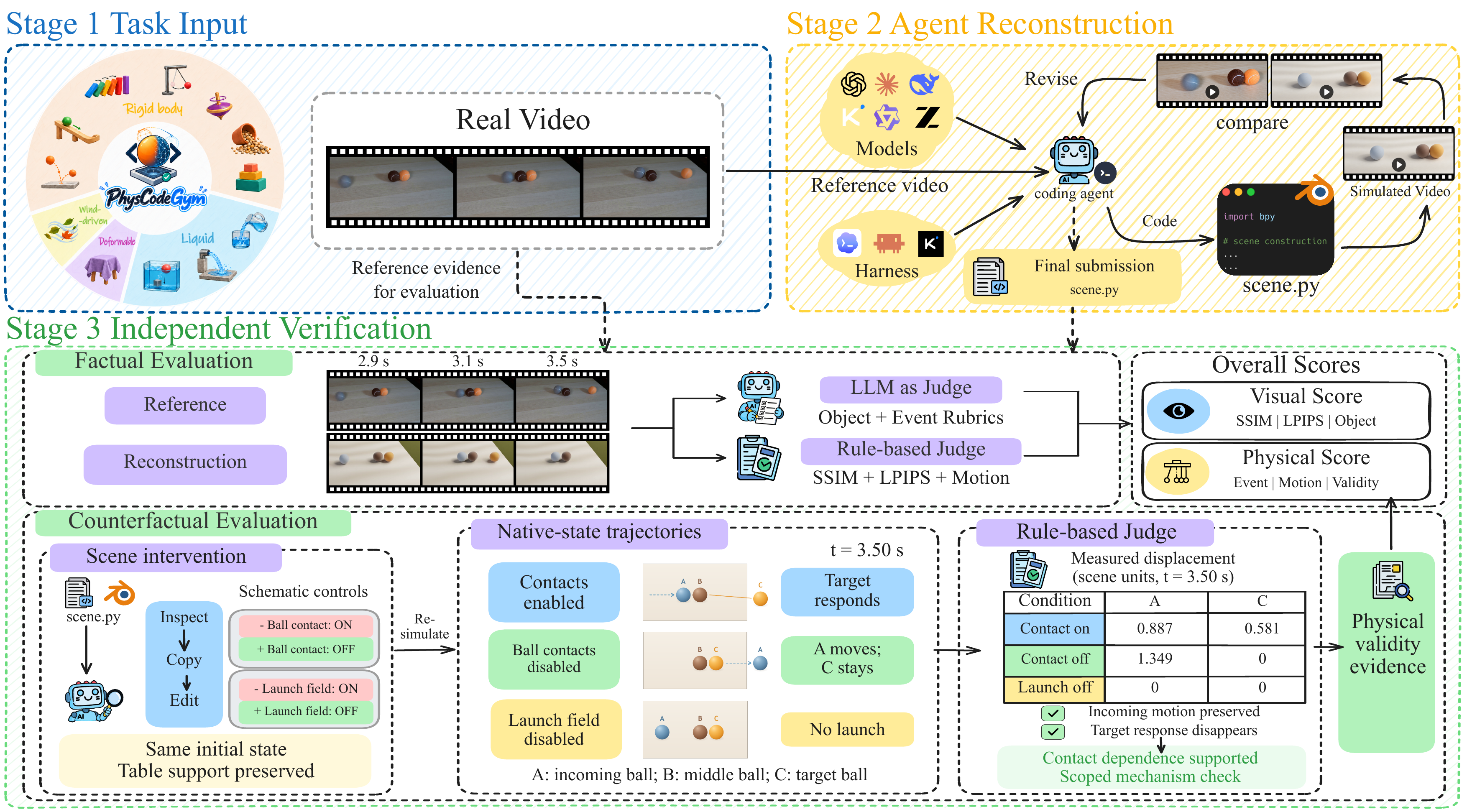}
  \caption{\textbf{Reconstruction and verification pipeline of PhysCodeGym.}
  \textbf{Stage 1:} A reference video defines the reconstruction task and supplies evidence for evaluation.
  \textbf{Stage 2:} A coding agent iterates between observing the video, writing a Blender scene program,
  running it, and comparing rendered previews, then submits the scene program.
  \textbf{Stage 3:} Independent evaluation combines rubric-based object and event judgments with
  image and motion metrics, while native-scene inspection and controlled re-simulation assess physical validity.
  In the illustrated contact test, disabling ball contacts preserves incoming motion but removes the target response;
  disabling the launch field removes the incoming motion as well.
  Visual and physical scores are reported separately.}
  \label{fig:overview}
\end{figure*}


\subsection{Dataset}
\label{sec:dataset}

The dataset contains $156$ tasks covering rigid-body, liquid, deformable, and wind-driven motion; detailed category definitions and task statistics are provided in \cref{sec:composition}.

\paragraph{Design principles.}
PhysCodeGym rests on four principles.
(P1) Visible physical causation: the motion in every reference clip is caused by a visible physical process whose cause can be inferred from the footage,
such as falling, rolling, collision, rotation, swinging, liquid pouring and splashing, deformable motion, and wind-driven motion.
(P2) Simulator expressibility: the process can be expressed within the simulation capabilities of Blender;
combustion, cutting, tearing and fracture, siphoning and absorption, liquid mixing, and magnetic forces are excluded.
(P3) Mechanism-level taxonomy: every task has one mutually exclusive main category,
and rigid-body and liquid tasks are further divided into subcategories by their dominant motion mechanism,
so that results can be analyzed per mechanism.
(P4) Observation-grounded annotation: annotations describe only what the reference video shows, with timestamped evidence;
what it does not show, such as the path of an object while it briefly leaves the frame, is not required.

\paragraph{Collection pipeline.}
Following (P1) and (P2), we collect candidate videos from Pexels~\citep{Free4KStock}, Physics-IQ~\citep{DBLP:conf/wacv/MotamedCSJG26},
BAAI-DataCube Physics-aware Videos~\citep{BAAIDataCubePhysicsawarevideosDatasets2026}, and IRIS~\citep{DBLP:conf/eccv/KhanbayovBSK26}.
We combine automated checks, visual agent review, and manual selection to identify clearly observable physical processes that can be reconstructed in Blender,
prioritizing stable cameras, simple backgrounds, unobstructed events and diverse physical phenomena.
From videos at least $8$ seconds long, we select a continuous $8$-second segment with pronounced motion
and standardize it to $3840\times2160$ resolution, $30$~FPS, and $240$ frames while preserving the original playback speed.
After deduplication and final manual review, we retain $156$ clips, each defining one task.

\paragraph{Annotation.}
Each task includes three structured annotations that specify the objects to reconstruct, the events to reproduce, and the physical constraints to satisfy.
(i) The object checklist describes object categories, counts, appearance attributes, and initial spatial layout.
(ii) Event annotations identify the participating objects and specify key actions, their temporal order, and the attainment and persistence of final states, supported by timestamped video evidence.
(iii) Physical requirements are recorded in manually annotated rubrics that define permitted initial conditions, actuation, and equivalent representations, together with criteria for checking the physical mechanisms of the reconstructed scene.
For selected tasks, counterfactual verification alters forces or contact conditions and re-simulates the scene to check whether its response is consistent with the task's physical requirements.
These annotations support object accuracy, event score, and physical validity, respectively (\cref{sec:evaluation}).

We draft object checklists with model assistance and event annotations from reference observations, then check both using reference videos and unrelated-video controls.

\subsection{Task Definition}
\label{sec:task}

Every task in PhysCodeGym poses the same problem, reconstructing the physical process recorded in a video,
and each submission is scored along two axes, each targeting a distinct capability:
visual reconstruction (reproducing what the reference video shows)
and physical reconstruction (reproducing the dynamic process through modeled physics).
\Cref{fig:overview} illustrates the reconstruction and verification pipeline.

\paragraph{Physical-process reconstruction.}
A video records only one presentation of a dynamic scene from one viewpoint.
To understand why the process unfolds as it does, or how it would change under a different viewpoint or a modified condition,
one needs not another similar video but a model of the physical process that can be run, inspected, and modified again.
Executable worlds support re-simulation under changed conditions and rendering from new viewpoints~\citep{wangCodeWorldsAgentic2026}.
PhysCodeGym therefore sets the reconstruction target to be a program.
Given a reference video, a coding agent must write a scene program
that starts from an empty Blender scene, builds geometry, materials, cameras, and lighting,
and sets the physical properties, initial conditions, constraints, and forces of the objects;
executing it yields a native scene that evolves over time, together with its rendered sequence.
The submitted reconstruction is the program and the native scene it produces;
the rendered sequence is only one presentation of this process from the reference viewpoint.
We choose Blender because it provides modeling, rigid-body, cloth, fluid, and force-field simulation, and rendering in a single environment~\citep{BlenderPythonAPI},
and because both the scene structure and the simulation state are accessible through its scripting interface,
so an agent's decisions about objects, contacts, and the causes of motion can be inspected directly in the scene.

\paragraph{Valid reconstruction.}
A valid reconstruction must satisfy two conditions:
(i) it reproduces the observations, matching the appearance, layout, and visible events of the reference video;
and (ii) the observed motion is produced by the modeled physical process.
The same footage can also be reproduced through keyframes, scripted poses, or hidden supports;
such a scene may match the frames, but it does not explain the motion and does not respond correctly when conditions change.
Prescribed trajectories can achieve high reconstruction scores while leaving physical mechanisms underconstrained~\citep{shen4DCodeBenchBenchmarkingAgents2026}.
The motion of core objects must therefore come from the physics solver, forces, constraints, and initial conditions.
This excludes scripted trajectories and force fields with no counterpart in the video,
such as an attractor that steers a ball along the observed path.
Explicitly modeled external actuators are allowed, provided that the objects they act on respond through physical interaction.
Physical parameters are not always identifiable from the observed motion: in a modeled free-fall and inclined-sliding scenario, friction can be inferred while mass remains unobservable~\citep{DBLP:conf/dagm/KandukuriA0S20}.
Monocular reconstruction also faces ambiguity in scale and unobserved geometry, for which physical dynamics can provide additional constraints~\citep{rhoMonoPhysicsEstimatingGeometry2026}.
PhysCodeGym therefore does not require recovering the true values of these parameters or the exact hidden structure, and accepts any implementation that reproduces the observations and satisfies the physical requirements of the task.

\paragraph{Observation and submission.}
The agent works in a terminal environment and uses command-line tools to observe the reference video and operate Blender.
The observation tools return basic video metadata, specified frames, frame grids, and clips sampled at a given rate.
Neither the source program nor the physical parameters of the reference scene are provided.
The build tools run the scene program, inspect the saved scene structure, and render single-frame previews of the simulation state.
The agent submits the scene program \texttt{scene.py}.
The program must be self-contained, build the entire scene procedurally, and bake the simulation cache before rendering.
It outputs the native scene file and a rendered sequence of $240$ frames at $30$~FPS and $640\times360$,
whose duration matches that of the reference video, $8$ seconds.
The program may not decode media, access the network, spawn subprocesses, or import modules dynamically,
which ensures that the scene is generated entirely by modeling statements.
Before submission, a checking tool verifies the format and policy requirements and reports any problems, so that the agent can fix them before the run ends.
The unified output specification allows every submission to be re-executed in an independent environment
and compared with the reference video frame by frame.

\paragraph{Scoring.}
We report separate visual and physical scores, their component metrics, and the overall score defined in \cref{sec:evaluation}.
The visual score measures how closely the rendered sequence resembles the reference video on screen,
and averages three visual metrics:
structural similarity (SSIM) and perceptual similarity (LPIPS) compare the frames at the pixel and perceptual levels,
and object accuracy checks the rendered objects against the object checklist.
The physical score measures the dynamic process of the scene,
and averages three physical metrics:
the event score checks whether the annotated events occur,
the motion score compares where and for how long motion occurs,
and physical validity checks the scene against the physical requirements.
Except for physical validity, which reads the program and the native scene, all metrics use only the rendered sequence.
Each metric is also reported individually, which separates failures that a single number would conflate~\citep{jainRigidBenchEvaluatingRigidBody2026},
such as a scene that matches the frames through scripted motion and a physically valid scene that misses the observed events.
Metric definitions are given in \cref{sec:evaluation}.

\section{Experiments}
\label{sec:experiments}


\subsection{Setup}
\label{sec:setup}

We build the execution framework of PhysCodeGym on Harbor~\citep{Harbor_Framework}, integrating our Blender reconstruction environment and benchmark-specific evaluation pipeline.
\Cref{sec:harbor-integration} describes this integration and its relationship to the agent harnesses used in our experiments.

All agents work in the terminal environment of \cref{sec:task}
and typically iterate between observing the reference, writing the program, running the simulation, comparing previews, and revising the scene.
Each launch of a Blender process counts as one iteration, and the iteration limit constitutes the generation budget;
observation commands do not count as iterations, but they consume running time and context, and the whole run is also bounded by a total time limit.
Unless stated otherwise, all agents use high reasoning effort, with a run-time limit of $3$ hours and at most $20$ Blender launches per task.
Each model is initially scheduled once per task.
Audited serving-infrastructure failures are handled by separately recorded recovery runs, preserving the original trials; verifier-side failures are repaired by re-executing archived programs.
Errors caused by the agent's own requests, such as exceeding the context window or sending a malformed image, count as failures of the agent.
A task that is run more than once is scored by its latest run that produced a rendered sequence, not by its best-scoring run;
a task for which no run produced a rendered sequence receives zero.

\begin{table*}[!t]
\centering
\caption{Visual and physical reconstruction scores on PhysCodeGym (higher is better). Metric definitions and aggregation follow \cref{sec:evaluation}. Best and second-best scores are \textbf{bold} and \underline{underlined}.}
\label{tab:results}
\small
\setlength{\tabcolsep}{3pt}
\begin{tabular*}{\textwidth}{@{\extracolsep{\fill}}llccccccccc@{}}
\toprule
\multirow{2}{*}{Harness} & \multirow{2}{*}{Model} & \multicolumn{4}{c}{Visual} & \multicolumn{4}{c}{Physical} & \multirow{2}{*}{$S_{\mathrm{tot}}{\uparrow}$} \\
\cmidrule(lr){3-6} \cmidrule(lr){7-10}
& & $S_{\mathrm{SSIM}}{\uparrow}$ & $S_{\mathrm{LPIPS}}{\uparrow}$ & $S_{\mathrm{obj}}{\uparrow}$ & $S_{\mathrm{vis}}{\uparrow}$ & $S_{\mathrm{evt}}{\uparrow}$ & $S_{\mathrm{mot}}{\uparrow}$ & $S_{\mathrm{val}}{\uparrow}$ & $S_{\mathrm{phys}}{\uparrow}$ & \\
\midrule
\multirow{4}{*}{Codex} & GPT-6-Astra & \textbf{0.610} & \textbf{0.561} & \textbf{0.909} & \textbf{0.693} & \textbf{0.805} & \textbf{0.324} & \textbf{0.768} & \textbf{0.633} & \textbf{0.663} \\
 & GPT-5.6-Sol & \underline{0.550} & \underline{0.448} & \underline{0.835} & \underline{0.611} & \underline{0.687} & \underline{0.210} & \underline{0.690} & \underline{0.529} & \underline{0.570} \\
 & GPT-6-Luna & 0.518 & 0.363 & 0.633 & 0.505 & 0.393 & 0.128 & 0.555 & 0.359 & 0.432 \\
 & GPT-5.6-Luna & 0.506 & 0.355 & 0.639 & 0.500 & 0.448 & 0.157 & 0.564 & 0.391 & 0.446 \\
\midrule
\multirow{4}{*}{Claude Code} & DeepSeek-V4.1-Flash & 0.514 & 0.398 & 0.615 & 0.509 & 0.446 & 0.200 & 0.522 & 0.391 & 0.450 \\
 & Claude Sonnet 5 & 0.513 & 0.315 & 0.481 & 0.436 & 0.368 & 0.111 & 0.559 & 0.347 & 0.392 \\
 & Qwen3.8-Flash & 0.354 & 0.239 & 0.410 & 0.334 & 0.307 & 0.090 & 0.350 & 0.252 & 0.293 \\
 & GLM-5.3-Flash & 0.286 & 0.226 & 0.362 & 0.291 & 0.240 & 0.103 & 0.295 & 0.213 & 0.252 \\
\midrule
Kimi Code & Kimi K2.8 & 0.435 & 0.304 & 0.477 & 0.405 & 0.349 & 0.110 & 0.448 & 0.302 & 0.354 \\
\bottomrule
\end{tabular*}
\end{table*}

\subsection{Models}
\label{sec:models}

We evaluate coding agents that autonomously write and run code in a terminal environment.
The main comparison in \cref{tab:results} uses three harnesses:
(i) Codex~\citep{CodexChatGPTAI} for GPT-5.6-Sol and GPT-5.6-Luna~\citep{GPT56FrontierIntelligence2026}, GPT-6-Luna~\citep{IntroducingGPT6Sol2026}, and GPT-6-Astra~\citep{GPT6AstraNew};
(ii) Claude Code~\citep{ClaudeCodeAnthropic} for Claude Sonnet 5~\citep{IntroducingClaudeSonnet2026}, GLM-5.3-Flash~\citep{GLM53FlashFrontierIntelligence}, DeepSeek-V4.1-Flash~\citep{IntroducingDeepSeekV41FlashSmarter2026}, and Qwen3.8-Flash~\citep{Qwen38flashAlibabaCloud};
and (iii) Kimi Code for Kimi K2.8 Preview, requested as \texttt{kimi-for-coding}~\citep{WhatsNewKimi}.
Prior work explores sampling, revision, and verification strategies for language agents~\citep{zhuScalingTesttimeCompute2025}, while studies of LLM reasoning show that the effectiveness of test-time computation depends on inference strategy and problem difficulty~\citep{DBLP:conf/iclr/Snell0XK25}.
We also compare DeepSeek-V4.1-Flash at low, high, and max configured reasoning effort in Claude Code, and at high effort in Codex.

\subsection{Evaluation}
\label{sec:evaluation}

Each submission is evaluated with three visual metrics, which compare the rendered sequence with the reference video,
and three physical metrics, which assess the dynamic process of the scene (\cref{sec:task}).

\paragraph{Visual metrics.}
Structural similarity is the per-frame SSIM~\citep{1284395} between the rendered sequence and the reference video.
Perceptual similarity is $\mathrm{clip}(1-d,0,1)$, where $d$ is the per-frame LPIPS~\citep{DBLP:conf/cvpr/ZhangIESW18} (AlexNet) distance.
Object accuracy summarizes a vision-language model's judgments against the object checklist as an aggregate pass rate.
Structural and perceptual similarity measure closeness at the pixel and perceptual levels, while object accuracy focuses on object-level correctness and is insensitive to subtle differences in lighting and texture.

\paragraph{Event score.}
A vision-language model judges the candidate video against the event annotation item by item
and cites time intervals in the candidate video as evidence.
When judging, the model sees still images of each role from the reference video, to confirm object identities,
together with the candidate video and supplementary temporal frames;
it does not receive the action descriptions of the reference video, and decides whether each event occurs based on the candidate video alone.
Each candidate is judged independently three times, and the majority verdict is taken per item.
The event score is the weighted fraction of passed items;
if a key event that defines the task fails, the score is capped at zero.
For example, if the ball does not roll around the tray, no credit is given even if the tray layout is correct.

\paragraph{Motion score.}
The motion score adopts the motion mask of Physics-IQ~\citep{DBLP:conf/wacv/MotamedCSJG26}:
grayscale frames are Gaussian-blurred and differenced against a running-average background,
then thresholded and morphologically processed to obtain the motion region of each frame.
We compare the reference and candidate masks with two IoU measures.
Spatial IoU compares the regions where motion occurs at any point in the video.
Weighted IoU also reflects how long motion lasts:
with $w_p$ and $\hat{w}_p$ the fractions of frames in which pixel $p$ is moving in the reference and the candidate,
it equals $\sum_p \min(w_p,\hat{w}_p) / \sum_p \max(w_p,\hat{w}_p)$.
The motion score is the average of the two.
This metric does not require recognizing objects, and complements motion details not covered by the event annotation,
such as the decay of swing amplitude or the extent of a liquid stream.

\paragraph{Physical validity.}
An agent verifies physical validity against manually annotated task rubrics.
It inspects the submitted program and native scene, executes Blender/Python measurements, and supports each judgment with evidence.
The main-table score $S_{\mathrm{val}}$ evaluates the submitted program and its Blender scene, counting each task once per configuration.
For selected tasks, \emph{counterfactual verification} intervenes on forces, contacts, or initial conditions and re-simulates the scene to test the physical dependencies specified by the rubric.
Counterfactual checks are scored as criteria of the physical-validity rubric.
The verification procedure and intervention example appear in \cref{sec:agent-verification}.

\paragraph{Score aggregation.}
We report visual, physical, and overall scores:
\begin{equation}
\begin{aligned}
S_{\mathrm{vis}} &= \tfrac{1}{3}\,(S_{\mathrm{SSIM}} + S_{\mathrm{LPIPS}} + S_{\mathrm{obj}}),\\
S_{\mathrm{phys}} &= \tfrac{1}{3}\,(S_{\mathrm{evt}} + S_{\mathrm{mot}} + S_{\mathrm{val}}),\\
S_{\mathrm{tot}} &= \tfrac{1}{2}\,(S_{\mathrm{vis}} + S_{\mathrm{phys}}).
\end{aligned}
\label{eq:scores}
\end{equation}
where $S_{\mathrm{SSIM}}$, $S_{\mathrm{LPIPS}}$, and $S_{\mathrm{obj}}$ are structural similarity, perceptual similarity, and object accuracy,
and $S_{\mathrm{evt}}$, $S_{\mathrm{mot}}$, and $S_{\mathrm{val}}$ are the event score, the motion score, and physical validity.
All three summaries are reported together with the component metrics.
For a fully measured task, $S_{\mathrm{tot}}$ is also the arithmetic mean of the six component metrics.
In model comparisons, a task for which no valid artifact is submitted receives zero on every metric and is counted in the denominator.
For the provisional validity results, scores use the lower endpoints of the rubric-score intervals; existing-artifact technical gaps are omitted while registered missing-artifact zeros remain included.
The physical aggregate is first computed per task and then averaged on exactly the same available-case subset as validity.
It therefore does not average the differently covered model-level columns of \cref{tab:results}.

For model-aggregate tables, $S_{\mathrm{tot}}$ directly averages the displayed visual and physical score columns.
The equivalence to a six-component mean over tasks requires all six metrics to share the same averaging set; it does not hold for differently covered columns merely by averaging their means.

\section{Results \& Analysis}
\label{sec:results}


We first present visual and physical reconstruction scores and resource use, then examine reasoning effort, harness settings, and cases exposing gaps between visual fidelity and physical correctness.

\subsection{Benchmark Results}
\label{sec:benchmark-results}

\Cref{tab:results} reports six metrics and aggregate scores for nine model--harness configurations at default high effort, following \cref{sec:evaluation}.

\paragraph{Overall performance.}
GPT-6-Astra leads in overall, visual, and physical scores ($0.663$, $0.693$, and $0.633$); GPT-5.6-Sol ranks second overall at $0.570$.

\paragraph{Performance across dimensions.}
All nine configurations score higher on objects than events; GPT-5.6-Sol scores $0.835$ and $0.687$, respectively.
Motion remains weak, with GPT-6-Astra reaching only $0.324$.
Rankings also vary: GPT-6-Luna has a slightly higher visual score but lower event and motion scores than GPT-5.6-Luna.

\begin{figure}[!t]
\centering
\includegraphics[width=\linewidth]{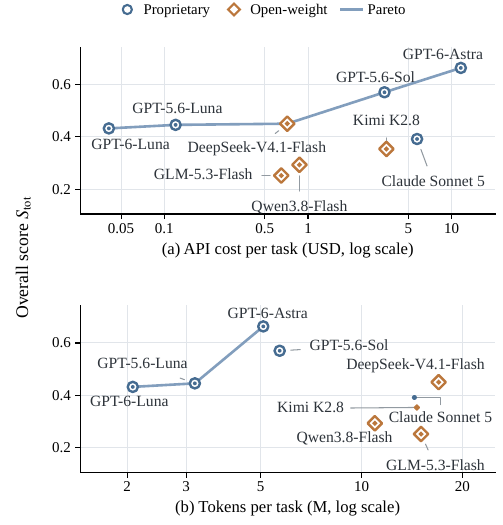}
\caption{Overall score $S_{\mathrm{tot}}$ versus (a) estimated API cost and (b) recorded token use per task. Greater cost or token use does not consistently yield better reconstruction across models.}
\label{fig:cost-quality}
\end{figure}

\paragraph{Resource use and quality.}
\Cref{fig:cost-quality} compares quality against estimated API cost and recorded token use.
GPT-5.6-Sol and GPT-6-Astra lie on the estimated-cost Pareto frontier: Sol scores $0.570$ at approximately \$3.40 per task, while Astra reaches $0.663$ at \$11.57, or $3.4$ times the cost.

Greater token use need not improve reconstruction.
DeepSeek-V4.1-Flash records $16.98$M cumulative tokens per task, nearly $3$ times GPT-5.6-Sol's $5.70$M, but scores $0.450$; its estimated cost is lower (\$0.72 per task).
Cumulative tokens include repeated context across interactions and cannot directly measure effective reasoning.
We next examine how effort and harness settings relate to budget use.

\subsection{Analysis}
\label{sec:results-analysis}

We analyze how artifact quality and completion shape aggregate scores through DeepSeek-V4.1-Flash effort and harness comparisons, followed by physical failure cases.

\paragraph{Reasoning effort and task completion.}
\Cref{fig:reasoning-effort} compares low, high, and max with the same limits of $3$ hours and $20$ Blender launches.
Low leads overall ($0.476$) when missing artifacts receive zero.
Among produced artifacts, max scores $0.547$ versus low's $0.527$, and also has the highest mean on the shared subset.
Gap decomposition links low's aggregate advantage mainly to completion, while max has longer runtimes and more recorded output tokens.
\Cref{sec:effort-analysis} details this completion--quality tradeoff under limited budgets.

\begingroup
\setlength{\intextsep}{6pt}
\begin{figure}[!ht]
\centering
\captionsetup{skip=3pt}
\includegraphics[width=\linewidth]{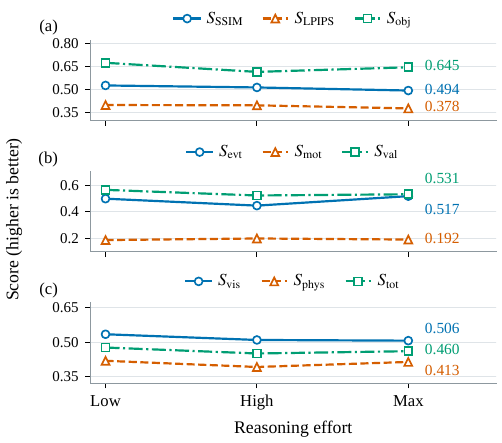}
\caption{Comparison of reasoning effort levels. We evaluate DeepSeek-V4.1-Flash on (a) visual metrics, (b) physical metrics, and (c) aggregate scores. Under the same budget of $3$ hours and $20$ Blender launches, low effort achieves the highest overall score, largely because it completes more tasks than max effort.}
\label{fig:reasoning-effort}
\end{figure}
\endgroup

\paragraph{Harness and verification.}
\Cref{fig:cross-harness} compares DeepSeek-V4.1-Flash at high effort in Claude Code and Codex.
Codex scores $0.479$ overall versus $0.450$, and $0.550$ on events versus $0.446$, with slightly lower motion overlap.
In the rolling-wood and wine-pouring traces, Codex identifies problems, revises, reruns, and inspects outputs; Claude Code's final revisions lack successful execution verification within the launch budget.
This pattern may relate to proactive verification guidance, check timing, and budget allocation (\cref{sec:cross-harness-cases}).

\begingroup
\setlength{\intextsep}{6pt}
\begin{figure}[!ht]
\centering
\captionsetup{skip=3pt}
\includegraphics[width=\linewidth]{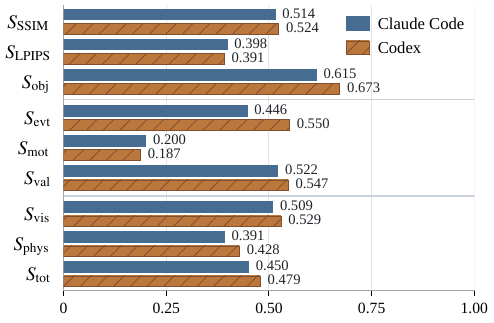}
\caption{Claude Code and Codex harness comparison. Visual, physical, and aggregate scores of DeepSeek-V4.1-Flash at high reasoning effort. In this comparison, Codex achieves a higher overall score, with the largest gain in event reconstruction despite slightly lower perceptual similarity and motion overlap.}
\label{fig:cross-harness}
\end{figure}
\endgroup

\paragraph{Physical failures beyond visual fidelity.}
Physical checks reveal flaws even when object and event judgments pass.
All six reconstructions in \cref{sec:validity-showcase} score $S_{\mathrm{obj}}=S_{\mathrm{evt}}=1$, yet program inspection and scene measurements expose failures such as unexplained energy increases and interpenetration.
Changing support, contact, or initial conditions further exposes latent defects beyond the original trajectories (\cref{sec:counterfactual-showcase}).

\section{Conclusion}
\label{sec:conclusion}

We introduced PhysCodeGym, a $156$-task benchmark for evaluating how coding agents reconstruct physical processes from video as executable Blender programs.
Visual and dynamic metrics assess rendered outcomes; verification agents inspect scene programs and perform measurements against task rubrics, including counterfactual rubric checks for selected tasks.
Our results highlight a gap between matching visible behavior and constructing physically valid scenes: object and event checks can pass while native-scene diagnostics identify physical inconsistencies.
The current benchmark is limited to processes that Blender can express and the physical checks specified by its rubrics.
Future work can extend PhysCodeGym to more simulation engines and cover more physical interactions.

{\fontsize{9}{10}\selectfont
\raggedbottom
\setlength{\bibsep}{1pt}
\bibliographystyle{ieeenat_fullname_url}
\bibliography{references}
\clearpage
}

\clearpage
\appendix
\crefalias{section}{appendix}
\onecolumn
\addtocontents{toc}{\protect\setcounter{tocdepth}{3}}
\begingroup
\makeatletter
\hypersetup{linkcolor=accentblue}
\pdfbookmark[0]{Appendix Contents}{appendix-contents}

{\centering
  \parbox{0.92\textwidth}{\centering
    \fontsize{18}{22}\selectfont\bfseries\appendixpapertitle}\par
  \vspace{10pt}
  {\fontsize{16}{20}\selectfont\normalfont Appendix\par}
}
\vspace{24pt}

\noindent{\fontsize{14}{17}\selectfont\bfseries Contents}\hfill
{\fontsize{9}{11}\selectfont Page}\par
\vspace{4pt}
\fontsize{10.95}{14.5}\selectfont

\let\appendix@contentsline\contentsline
\renewcommand{\contentsline}[4]{%
  \def\appendix@entry{#2}%
  \patchcmd{\appendix@entry}{\hskip -1em.~}{}{}{}%
  \appendix@contentsline{#1}{\appendix@entry}{#3}{#4}%
}
\renewcommand*{\l@subsection}{\@dottedtocline{2}{1.5em}{2.5em}}
\renewcommand*{\l@subsubsection}[2]{%
  {\fontsize{10}{13}\selectfont
   \@dottedtocline{3}{4.4em}{3.4em}{#1}{#2}}%
}
\@starttoc{toc}
\makeatother
\clearpage
\endgroup


\section{Dataset and Execution Framework}
\label{sec:dataset-appendix}

\subsection{Composition and Sources}
\label[appendix]{sec:composition}

The $156$ tasks contain $578$ event items in total (\cref{tab:data}).
By main category, there are $95$ rigid-body, $40$ liquid, $12$ deformable, and $9$ wind-driven tasks.

\begin{table}[!ht]
\centering
\caption{Composition of PhysCodeGym: number of tasks and event items per main category and subcategory.}
\label{tab:data}
\small
\setlength{\tabcolsep}{4pt}
\begin{tabular}{llrr}
\toprule
Category & Subcategory & Tasks & Event items \\
\midrule
\multirow{7}{*}{Rigid body} & Falling and bouncing & 16 & 71 \\
 & Rolling and tracks & 18 & 68 \\
 & Collision and chain reactions & 14 & 56 \\
 & Swinging and suspension & 16 & 51 \\
 & Rotation & 20 & 63 \\
 & Granular and multi-object piling & 7 & 22 \\
 & Stacking and static equilibrium & 4 & 13 \\
\midrule
\multirow{3}{*}{Liquid} & Pouring and container filling & 25 & 109 \\
 & Jets and surface flow & 8 & 25 \\
 & Liquid--object interaction & 7 & 30 \\
\midrule
Deformable & -- & 12 & 41 \\
Wind-driven & -- & 9 & 29 \\
\midrule
Total & & 156 & 578 \\
\bottomrule
\end{tabular}
\end{table}

\paragraph{Data sources.}
The reference videos come from Pexels~\citep{Free4KStock}, Physics-IQ~\citep{DBLP:conf/wacv/MotamedCSJG26},
BAAI-DataCube Physics-aware Videos~\citep{BAAIDataCubePhysicsawarevideosDatasets2026}, and IRIS~\citep{DBLP:conf/eccv/KhanbayovBSK26}.
\Cref{tab:data-sources} reports the number of retained tasks from each source by main category.

\begin{table}[!ht]
\centering
\caption{Data sources of PhysCodeGym. Each retained reference clip contributes one task.}
\label{tab:data-sources}
\small
\setlength{\tabcolsep}{7pt}
\begin{tabular}{lrrrrr}
\toprule
Source & Rigid body & Liquid & Deformable & Wind-driven & Total \\
\midrule
Pexels & 69 & 35 & 9 & 9 & 122 \\
Physics-IQ & 19 & 5 & 3 & 0 & 27 \\
BAAI-DataCube Physics-aware Videos & 5 & 0 & 0 & 0 & 5 \\
IRIS & 2 & 0 & 0 & 0 & 2 \\
\midrule
Total & 95 & 40 & 12 & 9 & 156 \\
\bottomrule
\end{tabular}
\end{table}

\paragraph{Rigid-body tasks.}
Rigid-body tasks fall into seven subcategories.
\emph{Falling and bouncing} tests gravity, collision restitution, and coming to rest after landing.
\emph{Rolling and tracks} covers objects rolling on planes, slopes, trays, or tracks, with paths determined by the track geometry.
\emph{Collision and chain reactions} focuses on momentum transfer between objects, as in billiards, bowling, dominoes, and Newton's cradles.
\emph{Swinging and suspension} involves swings, pendulums, metronomes, and hanging objects, whose motion is organized by hinges or ropes.
\emph{Rotation} involves turntables, windmills, spinning tops, and rotating platforms.
\emph{Granular and multi-object piling} involves pouring and piling of many objects, such as rice grains, flakes, and paper clips.
\emph{Stacking and static equilibrium} tests whether blocks or stones remain stable or topple under loading.

\paragraph{Liquid tasks.}
Liquid tasks fall into three subcategories.
\emph{Pouring and container filling} is the most common case: liquid flows from a bottle mouth, spout, or pipe into a container, and the liquid level rises accordingly.
\emph{Jets and surface flow} covers jets from hoses, nozzles, and fountains, as well as the spreading and splashing of liquid after it lands on a surface.
\emph{Liquid--object interaction} covers objects falling into liquid, sinking or floating in it, and liquid streams striking objects.

\paragraph{Deformable and wind-driven tasks.}
The deformable and wind-driven categories contain few tasks and are not subdivided.

\clearpage
\subsection{Reference Video Examples}
\label[appendix]{sec:dataset-examples}

\begin{figure}[!ht]
  \centering
  \includegraphics[width=\textwidth,height=0.82\textheight,keepaspectratio]{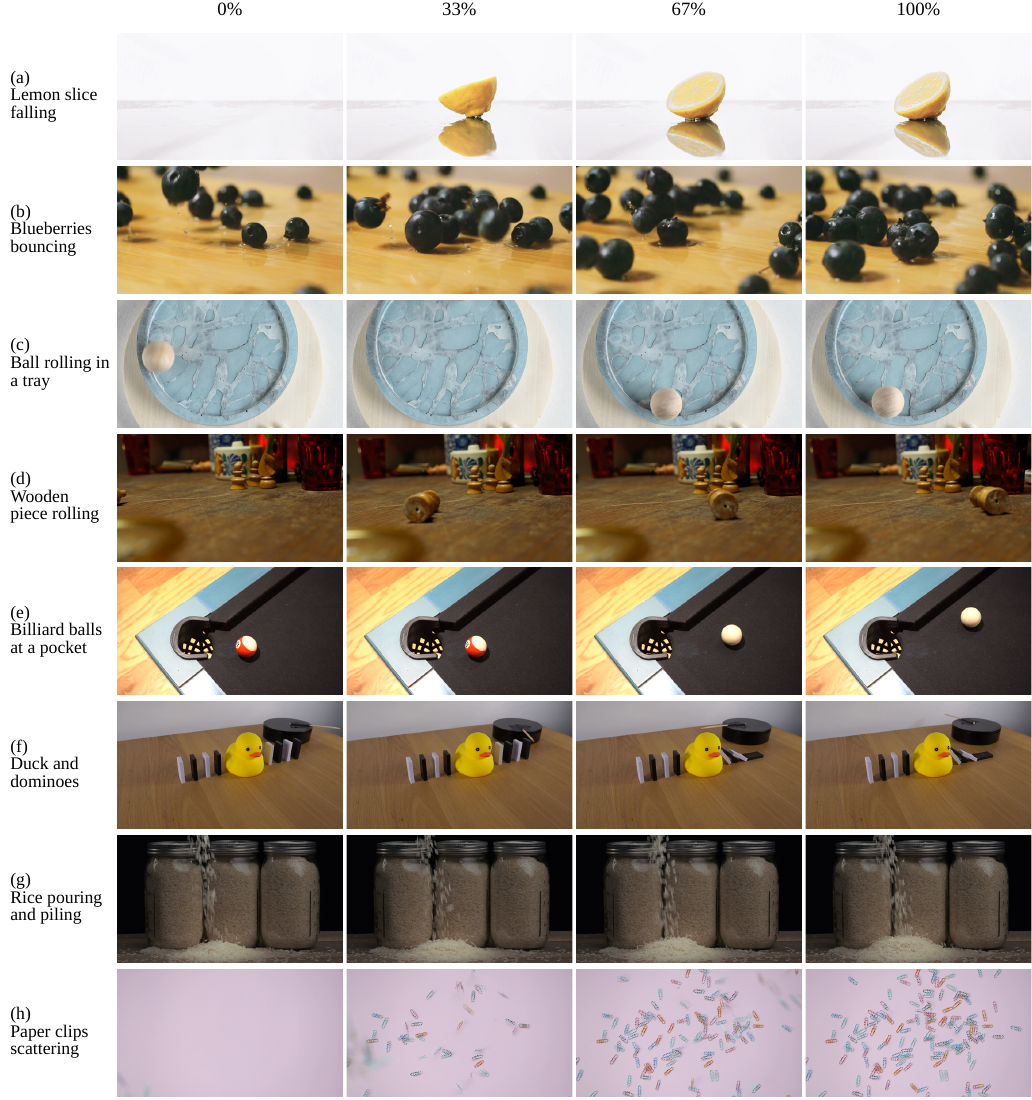}
  \caption{Rigid-body reference examples (1/2): falling and bouncing, rolling and
  tracks, collision and chain reactions, and granular or multi-object piling.
  Each row contains four uniformly sampled frames from one reference video.}
  \label{fig:dataset-rigid-a}
\end{figure}

\begin{figure}[!ht]
  \centering
  \includegraphics[width=\textwidth,height=0.82\textheight,keepaspectratio]{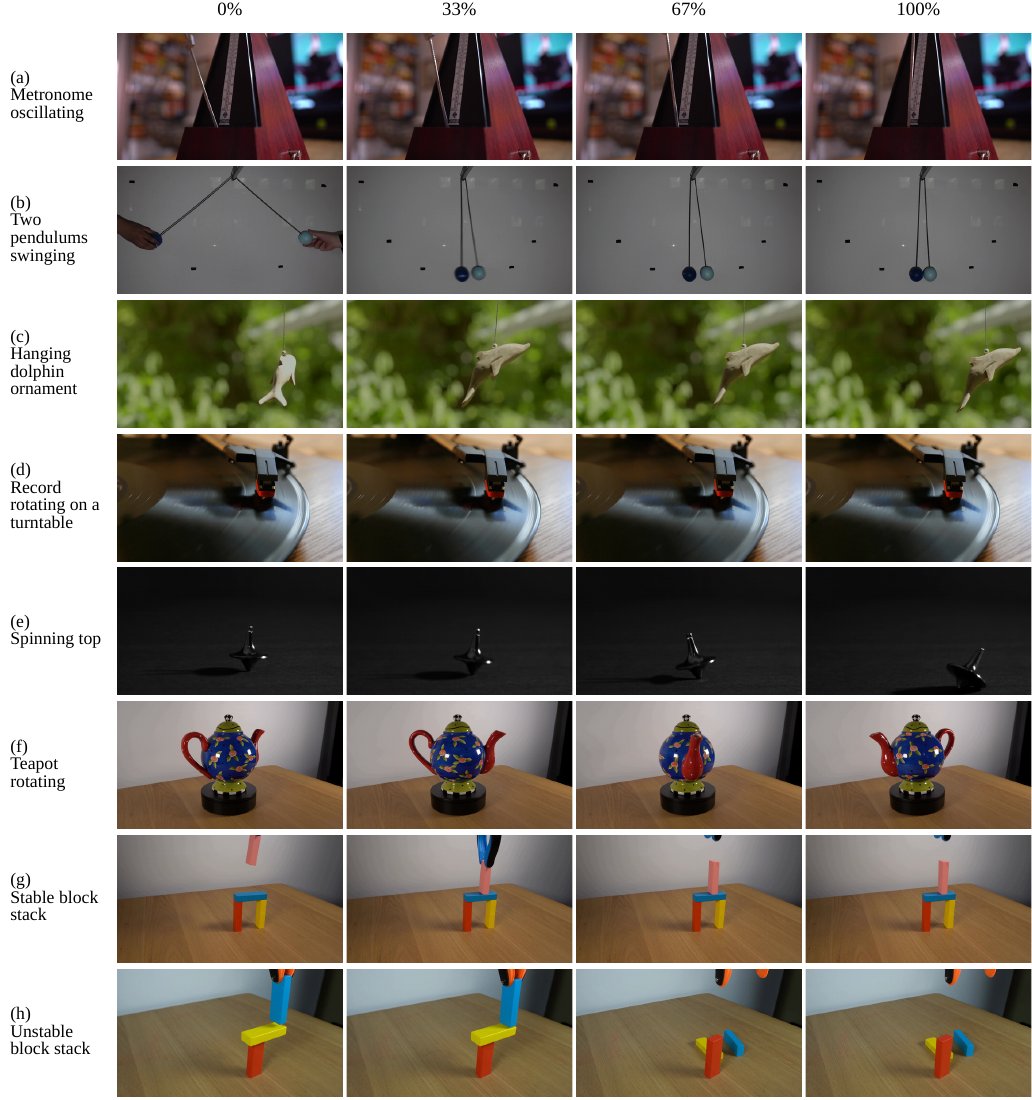}
  \caption{Rigid-body reference examples (2/2): swinging and suspension, rotation,
  and stacking or static equilibrium. Each row contains four uniformly sampled
  frames from one reference video.}
  \label{fig:dataset-rigid-b}
\end{figure}

\begin{figure}[!ht]
  \centering
  \includegraphics[width=\textwidth,height=0.82\textheight,keepaspectratio]{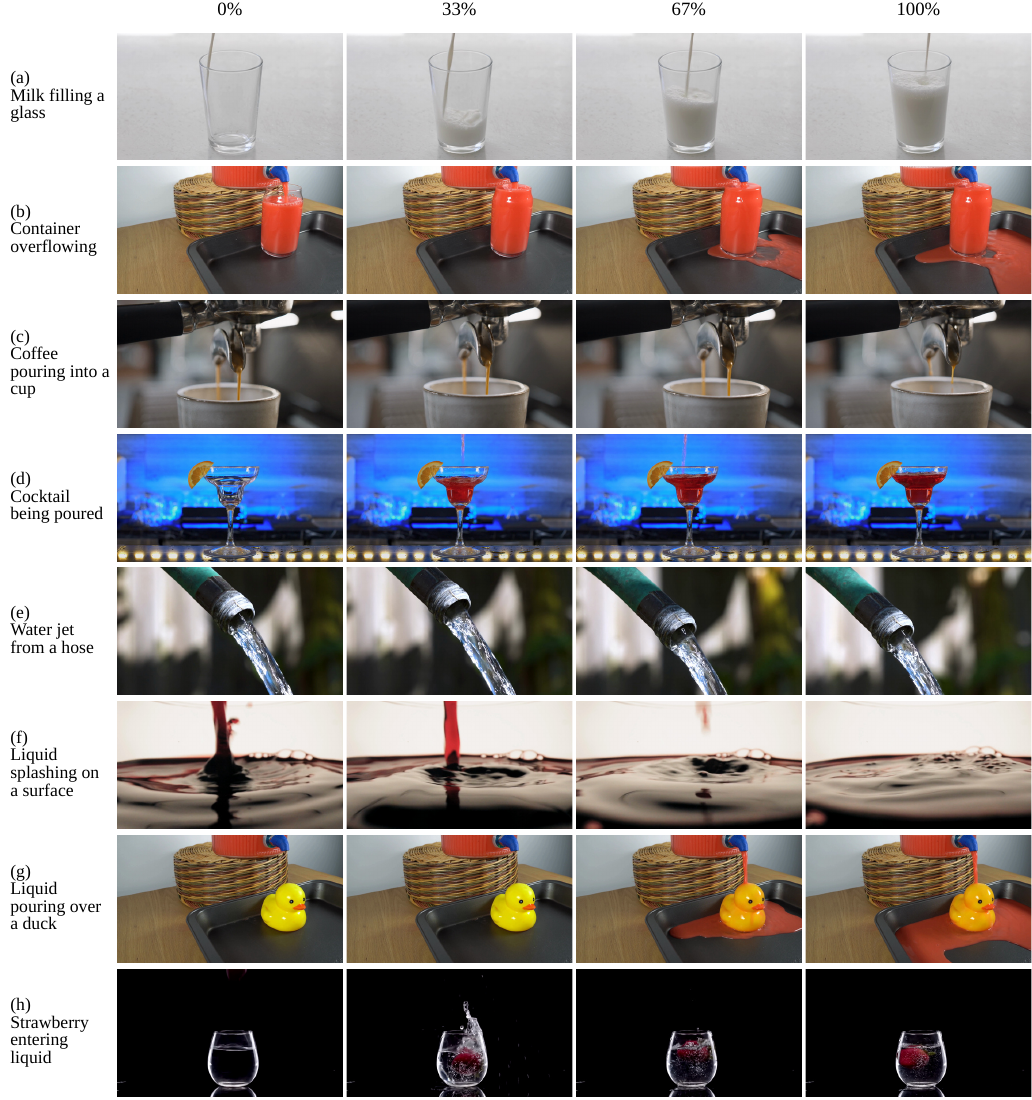}
  \caption{Liquid reference examples: pouring and container filling, jets and
  surface flow, and liquid--object interaction. Each row contains four uniformly
  sampled frames from one reference video.}
  \label{fig:dataset-liquid}
\end{figure}

\begin{figure}[!ht]
  \centering
  \includegraphics[width=\textwidth,height=0.82\textheight,keepaspectratio]{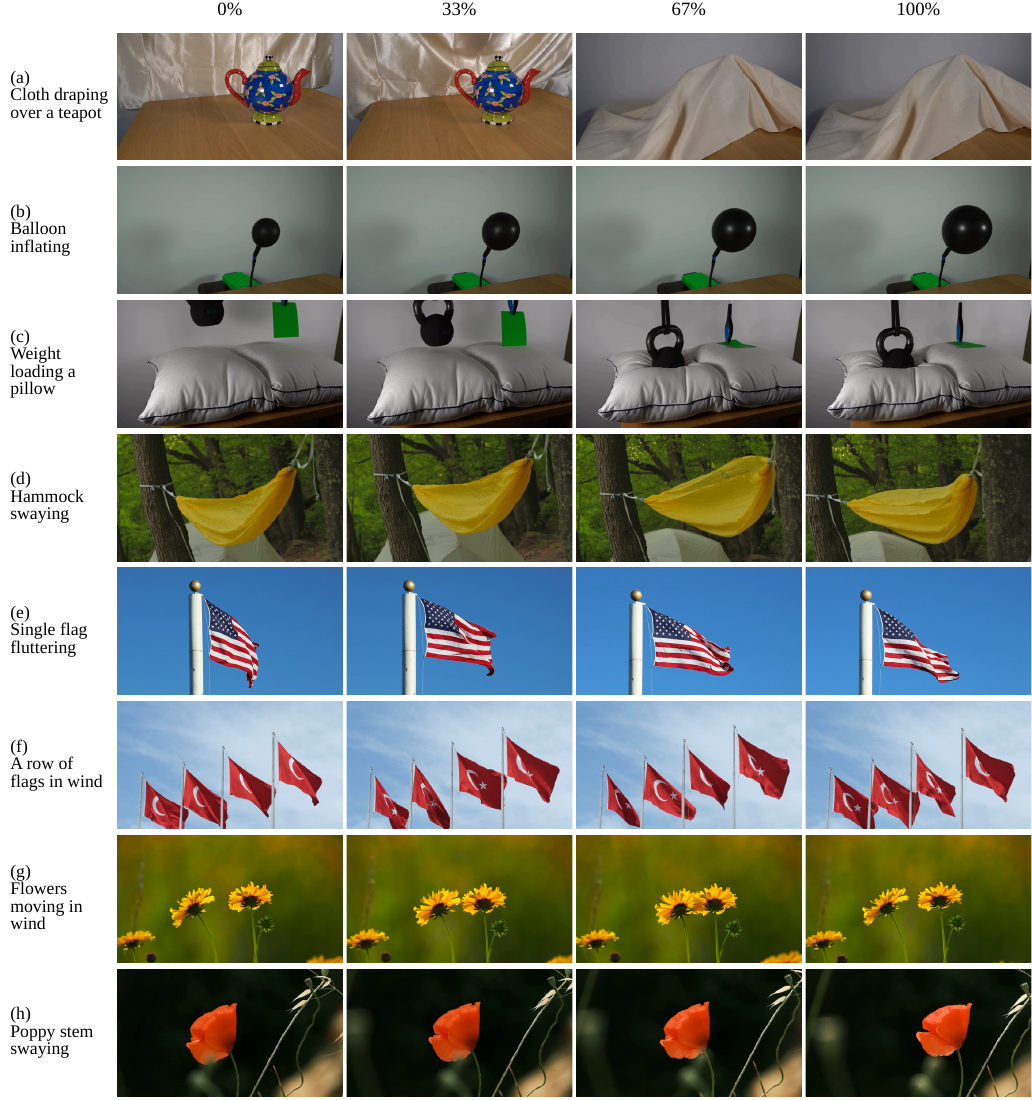}
  \caption{Deformable-category and wind-driven reference examples.
  Rows (a)--(d) show draping, membrane inflation, pillow deformation, and hammock
  motion; rows (e)--(h) show flag and plant motion in wind. Each row contains four
  uniformly sampled frames from one reference video.}
  \label{fig:dataset-cloth-wind}
\end{figure}

\Cref{fig:dataset-rigid-a,fig:dataset-rigid-b,fig:dataset-liquid,fig:dataset-cloth-wind}
show $32$ reference clips selected to illustrate the categories and subcategories
of PhysCodeGym. Each row shows four frames at approximately $0\%$, $33\%$, $67\%$,
and $100\%$ of the clip. Full frames are resized without cropping or color adjustment.
This selection illustrates the range of phenomena; the task frequencies are given
in \cref{sec:composition}.

\clearpage

\clearpage

\clearpage

\clearpage
\subsection{Harbor Integration}
\label[appendix]{sec:harbor-integration}

\paragraph{Framework organization.}
PhysCodeGym builds on Harbor~\citep{Harbor_Framework} v0.22.0, with local compatibility patches for CLI setup and container execution.
We use Harbor's Docker backend with Ubuntu 24.04 containers and the official Blender 4.0.2 distribution.
Harbor's installed-agent adapters launch the native Codex, Claude Code, and Kimi Code CLIs inside these containers (\cref{sec:models}).
The CLIs are installed in the image or mounted as preinstalled binaries; our Kimi adapter overrides only installation and inherits Harbor's Kimi Code execution path.
Harbor manages container and agent lifecycles, while PhysCodeGym supplies the reconstruction tasks, Blender tools, and evaluation protocol.

\paragraph{Reconstruction workflow.}
Each reconstruction task is packaged as a separate Harbor task containing task instructions, a reference video, and environment configuration.
The task image provides the Blender runtime and command-line tools for video observation, scene execution and inspection, and preview rendering.
The agent observes, programs, and checks its reconstruction inside the container (\cref{sec:task}).

Unless otherwise specified, Harbor limits agent execution to $3$ hours per run, and the task environment's launch interface permits at most $20$ Blender process launches (\cref{sec:setup}).
Scene execution, scene inspection, and preview rendering share this quota.
Video observation commands do not consume Blender launches but count toward execution time.
Once the launch quota is exhausted, new Blender launch requests are rejected; when the execution time limit is reached, Harbor ends the agent run.

\paragraph{Evaluation integration.}
After agent execution, Harbor collects the submitted program and invokes the benchmark's checking script in a separate verification container.
This stage executes the program, checks the scene and rendered outputs, and records basic image and motion measurements.
Subsequent evaluation jobs assess objects and events in the archived video.
A verification agent checks the program and scene against manually annotated physical criteria, including counterfactual tests for selected tasks (\cref{sec:evaluation-protocol}).
The component scores are aggregated according to the evaluation protocol in \cref{sec:evaluation}.

\clearpage
\section{Evaluation Protocol}
\label{sec:agent-verification}
\label{sec:evaluation-protocol}

This section describes task-level physical verification and counterfactual tests.

\subsection{Task Rubrics and Executable Verification}
\label[appendix]{sec:task-verification}

\paragraph{Task rubrics.}
We assess physical validity with an agent acting as a verifier, guided by manually annotated physical rubrics for each task.
Each rubric specifies the permitted initial setup and actuation, the mechanisms to check, and the evidence required for a judgment.
The criteria cover numerical stability, gravity, mass, and actuation, together with task-specific requirements on contact, support, and structural integrity.
The same rubric applies to every candidate for a task.

\paragraph{Executable verification.}
The verification agent reads the submitted program and native scene, maps the rubric's semantic roles to scene objects, and traces how their motion is implemented.
It formulates a measurement plan and writes probes adapted to that implementation.
The agent executes these probes in Blender to inspect geometry and measure object states, contact or support relations, and motion over time; Python tools analyze the resulting measurements.
It inspects the outputs and refines its measurements when further evidence is needed.
The final report gives a judgment for each criterion with supporting source references and measured evidence, and the evaluation harness applies the rubric's aggregation rules.
A missed reference event is assessed by the event score and is not by itself a physical violation.

\subsection{Counterfactual Tests}
\label[appendix]{sec:counterfactual-tests}

\emph{Counterfactual verification} evaluates rubric criteria that require controlled interventions.
The rubric specifies the physical dependence to test and the expected response under a controlled intervention.
The agent locates the corresponding mechanism in the program, creates an independent scene copy, implements the prescribed intervention, and re-simulates it for comparison with the unchanged baseline.
Interventions may alter a force, remove a contact, or change an initial condition while holding other relevant conditions fixed.
Before interpreting the response, the agent verifies that the intervention takes effect and uses control scenes to check the measurement.
For example, when a ball moves a target through collision, removing the ball--target contact before impact while preserving the ball's incoming motion should eliminate the target's impact-induced response (\cref{fig:overview}).
The agent compares the measured response with the rubric's expectation to determine whether the reconstructed motion depends on the required physical interaction.

\clearpage
\section{Additional Results and Analysis}
\label{sec:additional-results}

We compare scores on aligned task subsets, report quality conditional on producing an artifact, and examine the relationship between reasoning effort and reconstruction quality.

\subsection{Scores on Aligned Task Subsets}
\label[appendix]{sec:paired-results}
The main table's video metrics cover $156$ tasks, whereas physical scores exclude technical gaps.
\Cref{tab:validity-paired} recomputes all metrics on the task set used for each configuration's physical score, aligning the averaging set within each row.
Event and motion remain video measurements; validity uses the corresponding reports.
Task sets still differ across configurations, rather than forming a common intersection for all models.

\begin{table}[!ht]
\centering
\caption{Scores recomputed on each configuration's physical-evaluation task subset.}
\label{tab:validity-paired}
\small
\setlength{\tabcolsep}{0pt}
\renewcommand{\arraystretch}{1.0}
\begin{tabular*}{\textwidth}{@{}>{\raggedright\arraybackslash}p{105pt}@{\hspace{8pt}}>{\raggedright\arraybackslash}p{58pt}@{\hspace{8pt}}*{6}{>{\centering\arraybackslash}p{\dimexpr(\linewidth-179pt)/6\relax}}@{}}
\toprule
Model & Harness & $S_{\mathrm{vis}}{\uparrow}$ & $S_{\mathrm{evt}}{\uparrow}$ & $S_{\mathrm{mot}}{\uparrow}$ & $S_{\mathrm{val}}{\uparrow}$ & $S_{\mathrm{phys}}{\uparrow}$ & $S_{\mathrm{tot}}{\uparrow}$ \\
\midrule
GPT-6-Astra & Codex & \textbf{0.694} & \textbf{0.806} & \textbf{0.325} & \textbf{0.768} & \textbf{0.633} & \textbf{0.664} \\
GPT-5.6-Sol & Codex & \underline{0.611} & \underline{0.687} & \underline{0.210} & \underline{0.690} & \underline{0.529} & \underline{0.570} \\
GPT-6-Luna & Codex & 0.507 & 0.392 & 0.129 & 0.555 & 0.359 & 0.433 \\
GPT-5.6-Luna & Codex & 0.503 & 0.450 & 0.157 & 0.564 & 0.391 & 0.447 \\
DeepSeek-V4.1-Flash & Claude Code & 0.511 & 0.452 & 0.200 & 0.522 & 0.391 & 0.451 \\
Claude Sonnet 5 & Claude Code & 0.436 & 0.370 & 0.111 & 0.559 & 0.347 & 0.392 \\
Qwen3.8-Flash & Claude Code & 0.335 & 0.316 & 0.090 & 0.350 & 0.252 & 0.294 \\
GLM-5.3-Flash & Claude Code & 0.291 & 0.241 & 0.104 & 0.295 & 0.213 & 0.252 \\
Kimi K2.8 & Kimi Code & 0.404 & 0.348 & 0.110 & 0.448 & 0.302 & 0.353 \\
\bottomrule
\end{tabular*}
\par\vspace{3pt}\parbox{\textwidth}{\footnotesize All scores within a row use the same task subset. Best and second-best scores are \textbf{bold} and \underline{underlined}, respectively.}
\end{table}

For GPT-6-Astra, restricting the visual mean from $156$ to $155$ tasks changes $S_{\mathrm{vis}}$ from $0.693$ to $0.694$ and $S_{\mathrm{tot}}$ from $0.663$ to $0.664$, without changing individual task scores.

\subsection{Scores on Produced Artifacts}
\label[appendix]{sec:rendered-results}

\Cref{tab:rendered-results} averages the same selected artifacts as \cref{tab:results}, conditional on each configuration producing a rendered sequence.
Tasks without an output are excluded; zero scores on existing outputs are retained.
The five observed metrics use all rendered outputs, while validity and the physical score use those with an available physical judgment from either evidence scope, following \cref{sec:evaluation}.
$S_{\mathrm{tot}}$ averages the reported visual and physical scores, retaining their respective coverage.
These output-conditioned means describe reconstruction quality on produced artifacts; the full-task results also account for missing outputs.

\begin{table}[!ht]
\centering
\caption{Mean scores on tasks with rendered outputs for the configurations in \cref{tab:results}. Higher is better. Best and second-best scores are \textbf{bold} and \underline{underlined}.}
\label{tab:rendered-results}
\small
\setlength{\tabcolsep}{3pt}
\begin{tabular*}{\textwidth}{@{\extracolsep{\fill}}llccccccccc@{}}
\toprule
\multirow{2}{*}{Harness} & \multirow{2}{*}{Model} & \multicolumn{4}{c}{Visual} & \multicolumn{4}{c}{Physical} & \multirow{2}{*}{$S_{\mathrm{tot}}{\uparrow}$} \\
\cmidrule(lr){3-6} \cmidrule(lr){7-10}
& & $S_{\mathrm{SSIM}}{\uparrow}$ & $S_{\mathrm{LPIPS}}{\uparrow}$ & $S_{\mathrm{obj}}{\uparrow}$ & $S_{\mathrm{vis}}{\uparrow}$ & $S_{\mathrm{evt}}{\uparrow}$ & $S_{\mathrm{mot}}{\uparrow}$ & $S_{\mathrm{val}}{\uparrow}$ & $S_{\mathrm{phys}}{\uparrow}$ & \\
\midrule
\multirow{4}{*}{Codex} & GPT-6-Astra & \textbf{0.610} & \textbf{0.561} & \textbf{0.909} & \textbf{0.693} & \textbf{0.805} & \textbf{0.324} & \textbf{0.768} & \textbf{0.633} & \textbf{0.663} \\
 & GPT-5.6-Sol & 0.553 & 0.451 & \underline{0.840} & \underline{0.615} & \underline{0.692} & 0.211 & \underline{0.694} & \underline{0.532} & \underline{0.574} \\
 & GPT-6-Luna & 0.554 & 0.387 & 0.677 & 0.539 & 0.420 & 0.136 & 0.593 & 0.384 & 0.462 \\
 & GPT-5.6-Luna & 0.529 & 0.371 & 0.669 & 0.523 & 0.469 & 0.165 & 0.591 & 0.409 & 0.466 \\
\midrule
\multirow{4}{*}{Claude Code} & DeepSeek-V4.1-Flash & \underline{0.599} & \underline{0.463} & 0.716 & 0.593 & 0.520 & \underline{0.233} & 0.609 & 0.456 & 0.525 \\
 & Claude Sonnet 5 & 0.568 & 0.349 & 0.532 & 0.483 & 0.407 & 0.122 & 0.619 & 0.384 & 0.434 \\
 & Qwen3.8-Flash & 0.564 & 0.381 & 0.653 & 0.532 & 0.489 & 0.142 & 0.566 & 0.407 & 0.470 \\
 & GLM-5.3-Flash & 0.551 & 0.436 & 0.696 & 0.561 & 0.462 & 0.199 & 0.572 & 0.414 & 0.488 \\
\midrule
Kimi Code & Kimi K2.8 & 0.570 & 0.398 & 0.626 & 0.531 & 0.458 & 0.145 & 0.592 & 0.399 & 0.465 \\
\bottomrule
\end{tabular*}
\end{table}

\clearpage
\subsection{Reasoning Effort, Completion, and Reconstruction Quality}
\label[appendix]{sec:effort-analysis}

We separate the benchmark aggregate, which retains missing-output zeros, from reconstruction quality conditional on producing an artifact.
All comparisons use the same frozen scores and validity composition as the main results.
\Cref{tab:effort-score-scope} follows the score-aggregation convention in \cref{sec:evaluation}.

\begin{table}[!ht]
\centering
\caption{Overall scores under different averaging sets. Best and second-best scores are \textbf{bold} and \underline{underlined}.}
\label{tab:effort-score-scope}
\small
\begin{tabular*}{\textwidth}{@{\extracolsep{\fill}}lccc@{}}
\toprule
Averaging set & Low & High & Max \\
\midrule
Benchmark aggregate & \textbf{0.476} & 0.450 & \underline{0.460} \\
Each setting's produced artifacts & \underline{0.527} & 0.525 & \textbf{0.547} \\
Shared tasks with outputs and physical judgments & 0.533 & \underline{0.535} & \textbf{0.564} \\
\bottomrule
\end{tabular*}
\end{table}

\begin{figure}[!ht]
\centering
\includegraphics[width=\textwidth]{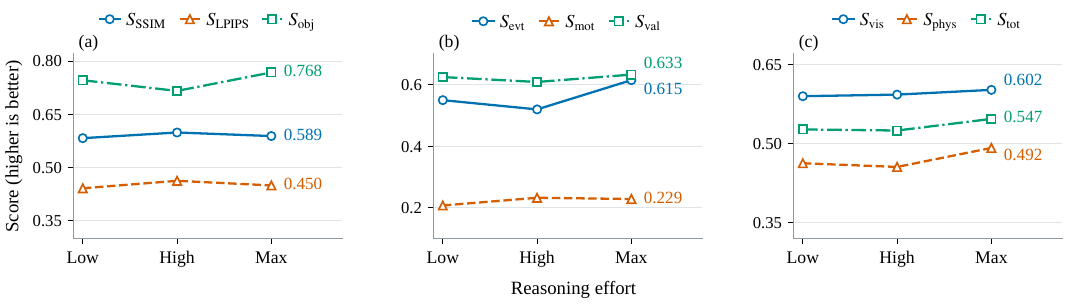}
\caption{DeepSeek-V4.1-Flash scores on tasks with rendered outputs (axis ranges differ).}
\label{fig:reasoning-effort-outputs}
\end{figure}

\paragraph{Quality conditional on an output.}
\Cref{fig:reasoning-effort-outputs} plots the nine scores using each setting's produced artifacts.
Video metrics use those artifacts, while validity and physical scores use the available physical judgments within that subset.
These subsets can differ between settings.
The overall ordering changes: max scores $0.547$, compared with $0.527$ for low and $0.525$ for high.
Max also has the highest conditional event and physical scores ($0.615$ and $0.492$).
Thus, the benchmark-level advantage of low does not imply better reconstruction quality whenever an output is produced.

\paragraph{Comparison on shared tasks.}
We also restrict all three settings to the intersection of tasks with both a submitted video and an available physical judgment for every setting, retaining zero scores on existing artifacts.
Max again has the highest visual, physical, and overall means on this shared subset.
Its event score is $0.628$, compared with $0.551$ for low and $0.535$ for high.
The mean paired, unrounded per-task overall-score difference between max and low is $0.030$; its paired task-bootstrap $95\%$ interval is $[-0.001,0.061]$.
The bootstrap measures sensitivity to task sampling; repeated-run variability is not estimated here.
This interval includes zero, so the observed mean ordering should not be interpreted as a universal effort advantage.

\paragraph{Where the aggregate gap arises.}
On the aligned task set with defined scores for both low and max, asymmetric missing outputs contribute $+0.036$ to the low-minus-max mean per-task score, while tasks with outputs from both settings contribute $-0.020$.
Their net contribution is $+0.016$.
This diagnostic uses the same eligible tasks and unrounded case scores, rather than recombining differently covered headline columns.
It identifies completion outcomes as the main source of low's aggregate lead.
Using original-condition-only validity reduces the low--max gap by approximately $0.003$, making the CF component a secondary contributor.

\paragraph{Execution overhead.}
Max uses $2.46\times$ the recorded output tokens of low on average, and mean agent execution time rises from $93.3$ to $110.7$ minutes under the fixed three-hour budget.
The additional missing outputs are concentrated in timeouts and final submissions that fail execution or lack required submission files.
These results highlight a tradeoff between task completion and reconstruction quality under a fixed execution budget.

\clearpage
\section{Case Studies}
\label{sec:case-studies}

These case studies connect evaluation scores with reconstructed behavior and executable evidence.
They cover metric interpretation, physical failures in unchanged scenes, counterfactual mechanism checks, and verification across harnesses.

\subsection{Metrics and Reconstruction Quality}
\label[appendix]{sec:metric-rankings}

Different metrics assess reconstruction appearance, event progression, and physical constraints.
The five cases in \cref{fig:diverse-pendulums,fig:diverse-rolling-egg,fig:diverse-wine,fig:diverse-avocado,fig:diverse-thai-blue}
examine differences in visible behavior and physical constraints that remain when individual metric scores are high or equal.
Each case pairs synchronized frames from a reference video and six reconstructions with component and summary scores.

\begin{table}[!ht]
\centering
\caption{Overview of the five case studies and their diagnostic questions.}
\label{tab:case-study-guide}
\small
\setlength{\tabcolsep}{5pt}
\renewcommand{\arraystretch}{1.22}
\begin{tabular*}{\textwidth}{@{}p{0.29\textwidth}@{\extracolsep{\fill}}p{0.22\textwidth}p{0.43\textwidth}@{}}
\toprule
Question & Cases & Main observation \\
\midrule
What differs when one metric is high? & Two pendulums; rolling quail egg & High motion overlap can coexist with incomplete events or physical violations. \\
\addlinespace[3pt]
What differs when one metric is equal? & Wine pouring; avocado halves & Equal event or validity scores coexist with differences in appearance and event progression. \\
\addlinespace[3pt]
What does scene inspection reveal? & Thai flag & An event score of $1$ coexists with cloth self-intersections. \\
\bottomrule
\end{tabular*}
\end{table}

\paragraph{Reading the figures and scores.}
Display times follow the reference event stages and are shared by the reference and all candidates; frames are shown without cropping.
Table rows follow the image-row order, identify each execution harness, and report six components together with the visual, physical, and overall scores.
The highest and second-highest distinct scores in each column are \textbf{bold} and \underline{underlined}, respectively; ties are determined before rounding.
Full videos and denser temporal samples accompany the cases.

Structural similarity, perceptual similarity, object accuracy, event score, and motion score evaluate submitted videos.
Physical validity evaluates the corresponding submitted program and its Blender scene.
All table metrics are higher-is-better ($\uparrow$); $S_{\mathrm{LPIPS}}$ is perceptual similarity, $\mathrm{clip}(1-d,0,1)$, rather than raw LPIPS distance $d$.
Score definitions and aggregation follow \cref{sec:evaluation}.
Component scores and their evidence identify the properties on which candidates differ.
All displayed score intervals are resolved, while validity judgments remain provisional pending semantic and evidence review.

\paragraph{Case and model selection.}
The archived 7 October 2026 pool contains $22$ eligible tasks, each with a GPT-6-Astra video and at least five other configurations with complete observed metrics and resolved physical-validity intervals.
Each task retains Astra and five other configurations selected by report availability, with configuration ID breaking ties; score values do not enter selection.
We inspect five non-flag candidates under randomized model labels and select four for distinct phenomena and visible differences, retaining one flag example.
All configurations use high effort, a three-hour budget, and at most twenty Blender launches.
Model sets can differ by task. These targeted cases explain evaluation dimensions; they do not estimate benchmark-wide ranking accuracy or constitute a human-preference experiment.

\clearpage
\subsubsection{Two Pendulums}
\begin{figure}[!ht]
\centering
\includegraphics[width=0.97\textwidth]{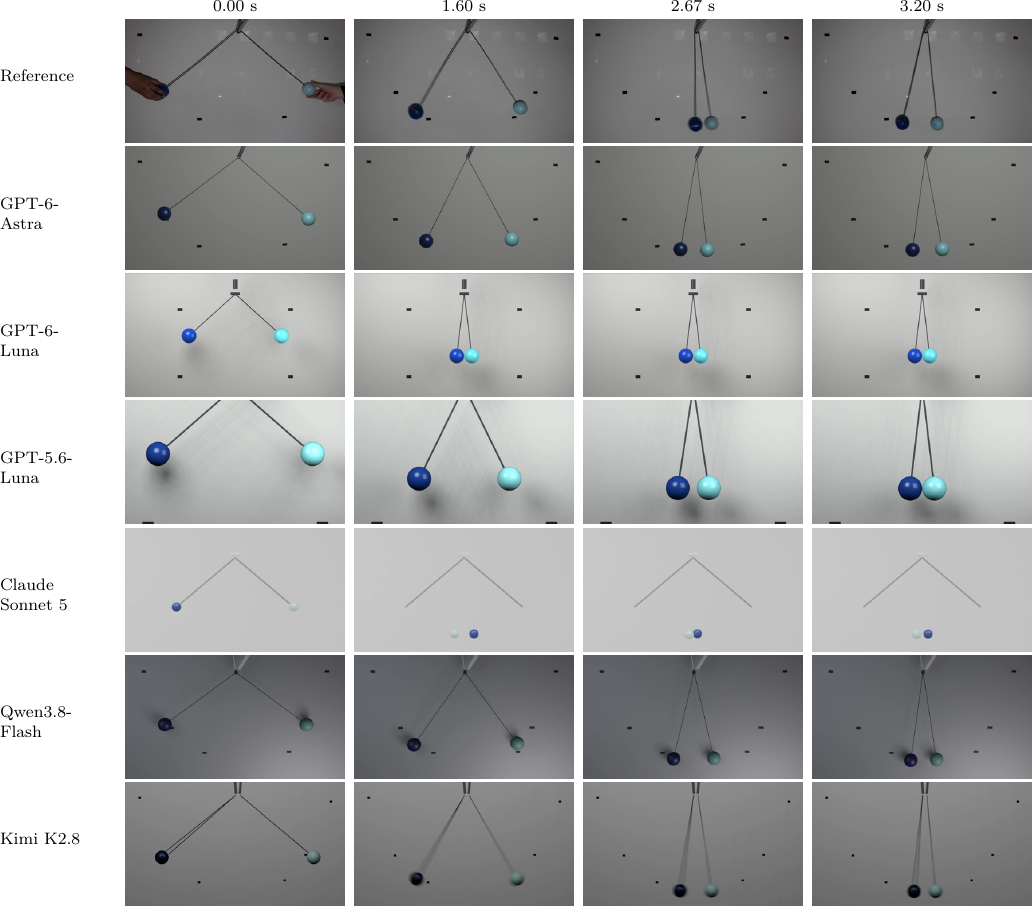}
\par\vspace{3pt}
\begingroup\fontsize{8}{9.5}\selectfont
\noindent\parbox{\linewidth}{\textbf{Component and summary scores.} Best and second-best scores are \textbf{bold} and \underline{underlined}, respectively.}\par\vspace{4pt}
\setlength{\tabcolsep}{0pt}
\renewcommand{\arraystretch}{1.18}
\begin{tabular*}{\linewidth}{@{}>{\raggedright\arraybackslash}p{76pt}@{\hspace{7pt}}>{\raggedright\arraybackslash}p{45pt}@{\hspace{7pt}}*{9}{>{\centering\arraybackslash}p{\dimexpr(\linewidth-135pt)/9\relax}}@{}}
\toprule
 & & \multicolumn{3}{c}{Visual metrics} & \multicolumn{3}{c}{Physical metrics} & \multicolumn{3}{c}{Summary scores} \\
\cmidrule(lr){3-5}\cmidrule(lr){6-8}\cmidrule(l){9-11}
{\fontsize{9}{10.5}\selectfont Model} & {\fontsize{9}{10.5}\selectfont Harness} & $S_{\rm SSIM}\!\uparrow$ & $S_{\rm LPIPS}\!\uparrow$ & $S_{\rm obj}\!\uparrow$ & $S_{\rm evt}\!\uparrow$ & $S_{\rm mot}\!\uparrow$ & $S_{\rm val}\!\uparrow$ & $S_{\rm vis}\!\uparrow$ & $S_{\rm phys}\!\uparrow$ & $S_{\rm tot}\!\uparrow$ \\
\midrule
GPT-6-Astra & Codex & \underline{0.905} & \textbf{0.801} & \textbf{1.000} & \textbf{1.000} & 0.242 & \textbf{1.000} & \textbf{0.902} & \underline{0.747} & \textbf{0.825} \\
GPT-6-Luna & Codex & 0.758 & 0.458 & \textbf{1.000} & 0.000 & 0.141 & \textbf{1.000} & 0.739 & 0.380 & 0.559 \\
GPT-5.6-Luna & Codex & 0.736 & 0.454 & \underline{0.857} & 0.000 & \textbf{0.318} & \underline{0.500} & 0.682 & 0.273 & 0.478 \\
Claude Sonnet 5 & Claude Code & 0.743 & 0.423 & \underline{0.857} & \underline{0.294} & 0.128 & \textbf{1.000} & 0.674 & 0.474 & 0.574 \\
Qwen3.8-Flash & Claude Code & \textbf{0.906} & 0.729 & \textbf{1.000} & \textbf{1.000} & \underline{0.306} & \underline{0.500} & 0.878 & 0.602 & 0.740 \\
Kimi K2.8 & Kimi Code & 0.891 & \underline{0.780} & \textbf{1.000} & \textbf{1.000} & 0.254 & \textbf{1.000} & \underline{0.891} & \textbf{0.751} & \underline{0.821} \\
\bottomrule
\end{tabular*}
\endgroup

\caption{\textbf{Motion overlap and event completion.} GPT-5.6-Luna has the highest motion IoU but an event score of $0$; GPT-6-Astra ranks fourth on motion IoU and scores $1$ on events. Luna's validity report identifies connector-solid overlap outside the shared pivot throughout frames 55--240, yielding validity $0.5$ versus Astra's $1$. Displayed frames cover early motion; scores cover all $8$ seconds.}
\label{fig:diverse-pendulums}
\end{figure}
\clearpage
\subsubsection{Rolling Quail Egg}
\begin{figure}[!ht]
\centering
\includegraphics[width=0.97\textwidth]{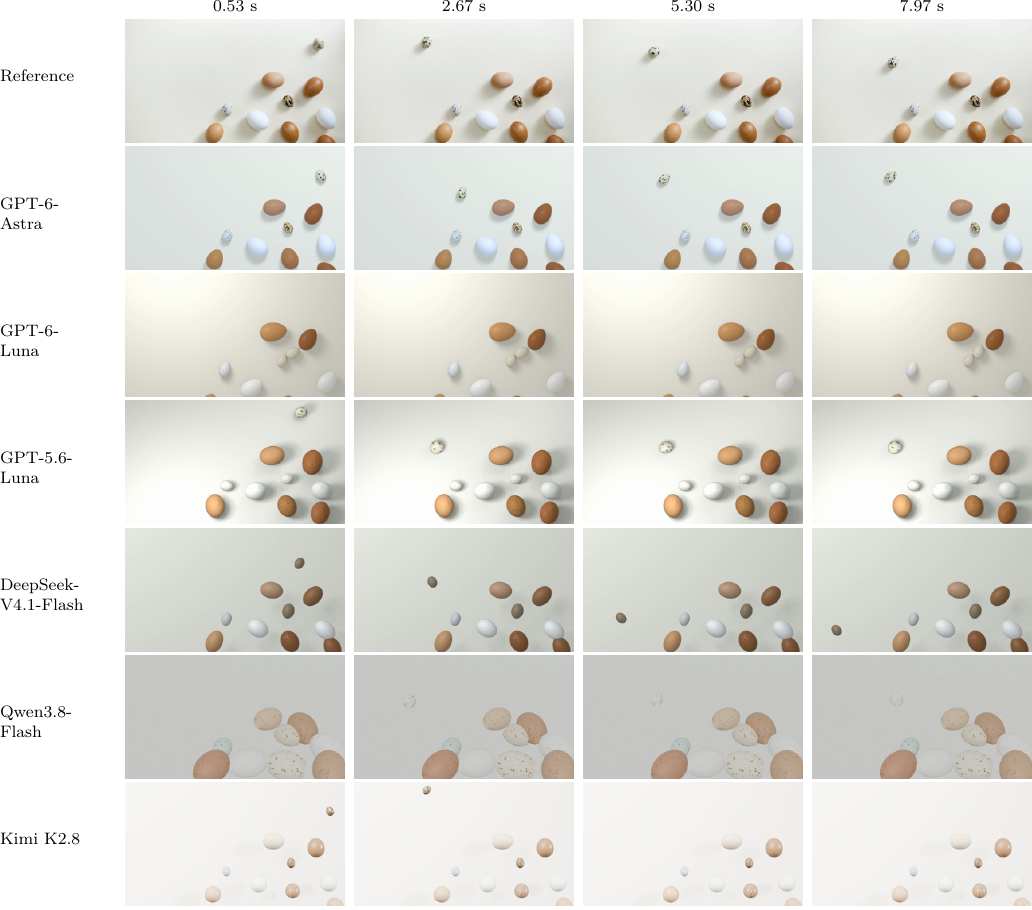}
\par\vspace{3pt}
\begingroup\fontsize{8}{9.5}\selectfont
\noindent\parbox{\linewidth}{\textbf{Component and summary scores.} Best and second-best scores are \textbf{bold} and \underline{underlined}, respectively.}\par\vspace{4pt}
\setlength{\tabcolsep}{0pt}
\renewcommand{\arraystretch}{1.18}
\begin{tabular*}{\linewidth}{@{}>{\raggedright\arraybackslash}p{76pt}@{\hspace{7pt}}>{\raggedright\arraybackslash}p{45pt}@{\hspace{7pt}}*{9}{>{\centering\arraybackslash}p{\dimexpr(\linewidth-135pt)/9\relax}}@{}}
\toprule
 & & \multicolumn{3}{c}{Visual metrics} & \multicolumn{3}{c}{Physical metrics} & \multicolumn{3}{c}{Summary scores} \\
\cmidrule(lr){3-5}\cmidrule(lr){6-8}\cmidrule(l){9-11}
{\fontsize{9}{10.5}\selectfont Model} & {\fontsize{9}{10.5}\selectfont Harness} & $S_{\rm SSIM}\!\uparrow$ & $S_{\rm LPIPS}\!\uparrow$ & $S_{\rm obj}\!\uparrow$ & $S_{\rm evt}\!\uparrow$ & $S_{\rm mot}\!\uparrow$ & $S_{\rm val}\!\uparrow$ & $S_{\rm vis}\!\uparrow$ & $S_{\rm phys}\!\uparrow$ & $S_{\rm tot}\!\uparrow$ \\
\midrule
GPT-6-Astra & Codex & \textbf{0.919} & \textbf{0.864} & \textbf{1.000} & \textbf{1.000} & \underline{0.225} & \textbf{1.000} & \textbf{0.927} & \textbf{0.742} & \textbf{0.835} \\
GPT-6-Luna & Codex & 0.895 & 0.753 & 0.500 & 0.000 & 0.000 & 0.450 & 0.716 & 0.150 & 0.433 \\
GPT-5.6-Luna & Codex & 0.851 & 0.708 & \textbf{1.000} & \textbf{1.000} & 0.083 & 0.450 & \underline{0.853} & 0.511 & 0.682 \\
DeepSeek-V4.1-Flash & Claude Code & \underline{0.907} & \underline{0.822} & 0.500 & \textbf{1.000} & \textbf{0.235} & \underline{0.500} & 0.743 & 0.578 & 0.660 \\
Qwen3.8-Flash & Claude Code & 0.873 & 0.705 & 0.333 & \textbf{1.000} & 0.001 & 0.450 & 0.637 & 0.484 & 0.560 \\
Kimi K2.8 & Kimi Code & 0.896 & 0.796 & \underline{0.833} & \underline{0.625} & 0.124 & \textbf{1.000} & 0.842 & \underline{0.583} & \underline{0.712} \\
\bottomrule
\end{tabular*}
\endgroup

\caption{\textbf{Motion overlap and scene constraints.} DeepSeek-V4.1-Flash exceeds GPT-6-Astra on motion IoU ($0.235$ versus $0.225$), but its validity is capped at $0.5$, versus Astra's $1$. The validity report identifies persistent interpenetration between two stationary eggs across all $240$ frames; this violation concerns the resting cluster rather than the moving egg. Astra retains the cluster and speckled appearance, with a timing mismatch.}
\label{fig:diverse-rolling-egg}
\end{figure}
\clearpage
\subsubsection{Wine Pouring}
\begin{figure}[!ht]
\centering
\includegraphics[width=0.97\textwidth]{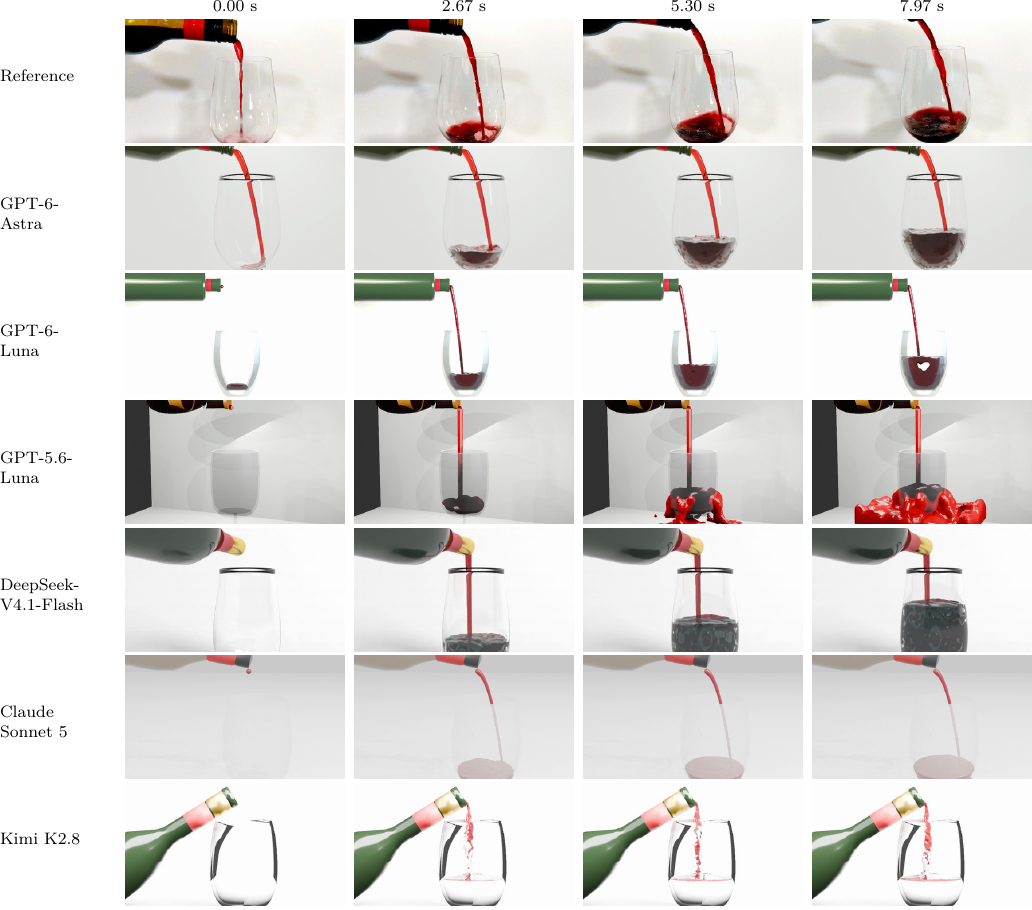}
\par\vspace{3pt}
\begingroup\fontsize{8}{9.5}\selectfont
\noindent\parbox{\linewidth}{\textbf{Component and summary scores.} Best and second-best scores are \textbf{bold} and \underline{underlined}, respectively.}\par\vspace{4pt}
\setlength{\tabcolsep}{0pt}
\renewcommand{\arraystretch}{1.18}
\begin{tabular*}{\linewidth}{@{}>{\raggedright\arraybackslash}p{76pt}@{\hspace{7pt}}>{\raggedright\arraybackslash}p{45pt}@{\hspace{7pt}}*{9}{>{\centering\arraybackslash}p{\dimexpr(\linewidth-135pt)/9\relax}}@{}}
\toprule
 & & \multicolumn{3}{c}{Visual metrics} & \multicolumn{3}{c}{Physical metrics} & \multicolumn{3}{c}{Summary scores} \\
\cmidrule(lr){3-5}\cmidrule(lr){6-8}\cmidrule(l){9-11}
{\fontsize{9}{10.5}\selectfont Model} & {\fontsize{9}{10.5}\selectfont Harness} & $S_{\rm SSIM}\!\uparrow$ & $S_{\rm LPIPS}\!\uparrow$ & $S_{\rm obj}\!\uparrow$ & $S_{\rm evt}\!\uparrow$ & $S_{\rm mot}\!\uparrow$ & $S_{\rm val}\!\uparrow$ & $S_{\rm vis}\!\uparrow$ & $S_{\rm phys}\!\uparrow$ & $S_{\rm tot}\!\uparrow$ \\
\midrule
GPT-6-Astra & Codex & \underline{0.766} & \textbf{0.589} & \textbf{1.000} & \textbf{1.000} & 0.200 & \textbf{1.000} & \textbf{0.785} & \underline{0.733} & \textbf{0.759} \\
GPT-6-Luna & Codex & 0.741 & 0.491 & 0.633 & \textbf{1.000} & 0.107 & \underline{0.500} & 0.622 & 0.536 & 0.579 \\
GPT-5.6-Luna & Codex & 0.697 & 0.457 & \underline{0.800} & \textbf{1.000} & \underline{0.209} & 0.450 & 0.651 & 0.553 & 0.602 \\
DeepSeek-V4.1-Flash & Claude Code & 0.718 & \underline{0.519} & \underline{0.800} & \textbf{1.000} & \textbf{0.314} & \textbf{1.000} & \underline{0.679} & \textbf{0.771} & \underline{0.725} \\
Claude Sonnet 5 & Claude Code & \textbf{0.774} & 0.511 & 0.533 & \textbf{1.000} & 0.056 & \underline{0.500} & 0.606 & 0.519 & 0.562 \\
Kimi K2.8 & Kimi Code & 0.685 & 0.442 & \underline{0.800} & \textbf{1.000} & 0.097 & \textbf{1.000} & 0.642 & 0.699 & 0.671 \\
\bottomrule
\end{tabular*}
\endgroup

\caption{\textbf{Appearance differences under equal event scores.} All six candidates receive an event score of $1$, despite differences in stream color, glass visibility, and accumulated liquid. Claude Sonnet 5 has the highest SSIM ($0.774$) despite its pale rendering, but lower perceptual similarity than Astra ($0.511$ versus $0.589$). Equal event scores thus coexist with appearance differences reflected in the frames and visual metrics.}
\label{fig:diverse-wine}
\end{figure}
\clearpage
\subsubsection{Avocado Halves Entering Water}
\begin{figure}[!ht]
\centering
\includegraphics[width=0.97\textwidth]{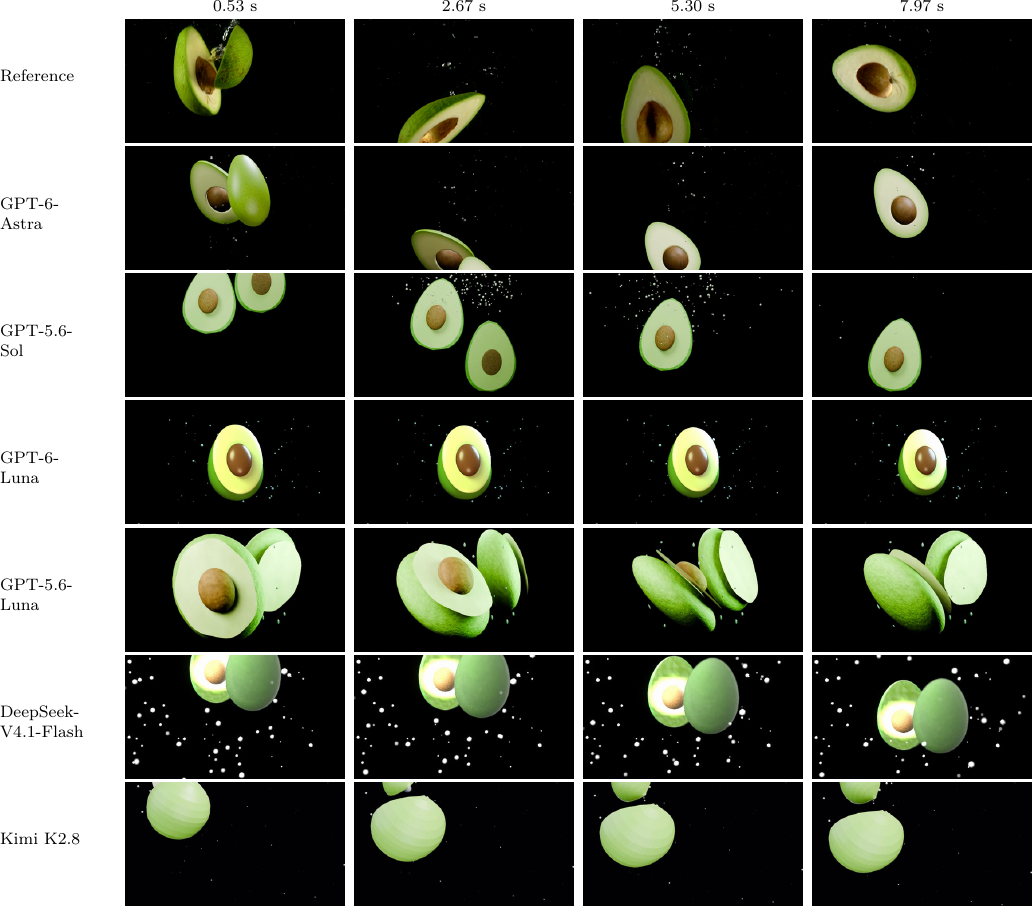}
\par\vspace{3pt}
\begingroup\fontsize{8}{9.5}\selectfont
\noindent\parbox{\linewidth}{\textbf{Component and summary scores.} Best and second-best scores are \textbf{bold} and \underline{underlined}, respectively.}\par\vspace{4pt}
\setlength{\tabcolsep}{0pt}
\renewcommand{\arraystretch}{1.18}
\begin{tabular*}{\linewidth}{@{}>{\raggedright\arraybackslash}p{76pt}@{\hspace{7pt}}>{\raggedright\arraybackslash}p{45pt}@{\hspace{7pt}}*{9}{>{\centering\arraybackslash}p{\dimexpr(\linewidth-135pt)/9\relax}}@{}}
\toprule
 & & \multicolumn{3}{c}{Visual metrics} & \multicolumn{3}{c}{Physical metrics} & \multicolumn{3}{c}{Summary scores} \\
\cmidrule(lr){3-5}\cmidrule(lr){6-8}\cmidrule(l){9-11}
{\fontsize{9}{10.5}\selectfont Model} & {\fontsize{9}{10.5}\selectfont Harness} & $S_{\rm SSIM}\!\uparrow$ & $S_{\rm LPIPS}\!\uparrow$ & $S_{\rm obj}\!\uparrow$ & $S_{\rm evt}\!\uparrow$ & $S_{\rm mot}\!\uparrow$ & $S_{\rm val}\!\uparrow$ & $S_{\rm vis}\!\uparrow$ & $S_{\rm phys}\!\uparrow$ & $S_{\rm tot}\!\uparrow$ \\
\midrule
GPT-6-Astra & Codex & \textbf{0.771} & \textbf{0.563} & \textbf{1.000} & \textbf{1.000} & \textbf{0.516} & \textbf{1.000} & \textbf{0.778} & \textbf{0.839} & \textbf{0.808} \\
GPT-5.6-Sol & Codex & \underline{0.702} & \underline{0.519} & 0.750 & \textbf{1.000} & \underline{0.339} & \textbf{1.000} & \underline{0.657} & \underline{0.780} & \underline{0.718} \\
GPT-6-Luna & Codex & 0.652 & 0.480 & 0.125 & 0.000 & 0.000 & \textbf{1.000} & 0.419 & 0.333 & 0.376 \\
GPT-5.6-Luna & Codex & 0.580 & 0.407 & 0.250 & 0.000 & 0.332 & \underline{0.500} & 0.412 & 0.277 & 0.345 \\
DeepSeek-V4.1-Flash & Claude Code & 0.573 & 0.383 & \underline{0.875} & 0.000 & 0.269 & \underline{0.500} & 0.610 & 0.256 & 0.433 \\
Kimi K2.8 & Kimi Code & 0.405 & 0.509 & 0.000 & \underline{0.471} & 0.191 & 0.150 & 0.305 & 0.270 & 0.288 \\
\bottomrule
\end{tabular*}
\endgroup

\caption{\textbf{Equal validity with different reconstruction fidelity.} GPT-6-Astra, GPT-5.6-Sol and GPT-6-Luna receive validity $1$, but event scores of $1$, $1$, and $0$, respectively. Luna's validity report accepts a single composite rigid body in a homogeneous analytical medium: measured motion responds to effective gravity and linear drag. Passing these checks under the permitted simplification does not establish reproduction of the reference events.}
\label{fig:diverse-avocado}
\end{figure}
\clearpage
\subsubsection{Thai Flag}
\begin{figure}[!ht]
\centering
\includegraphics[width=0.97\textwidth]{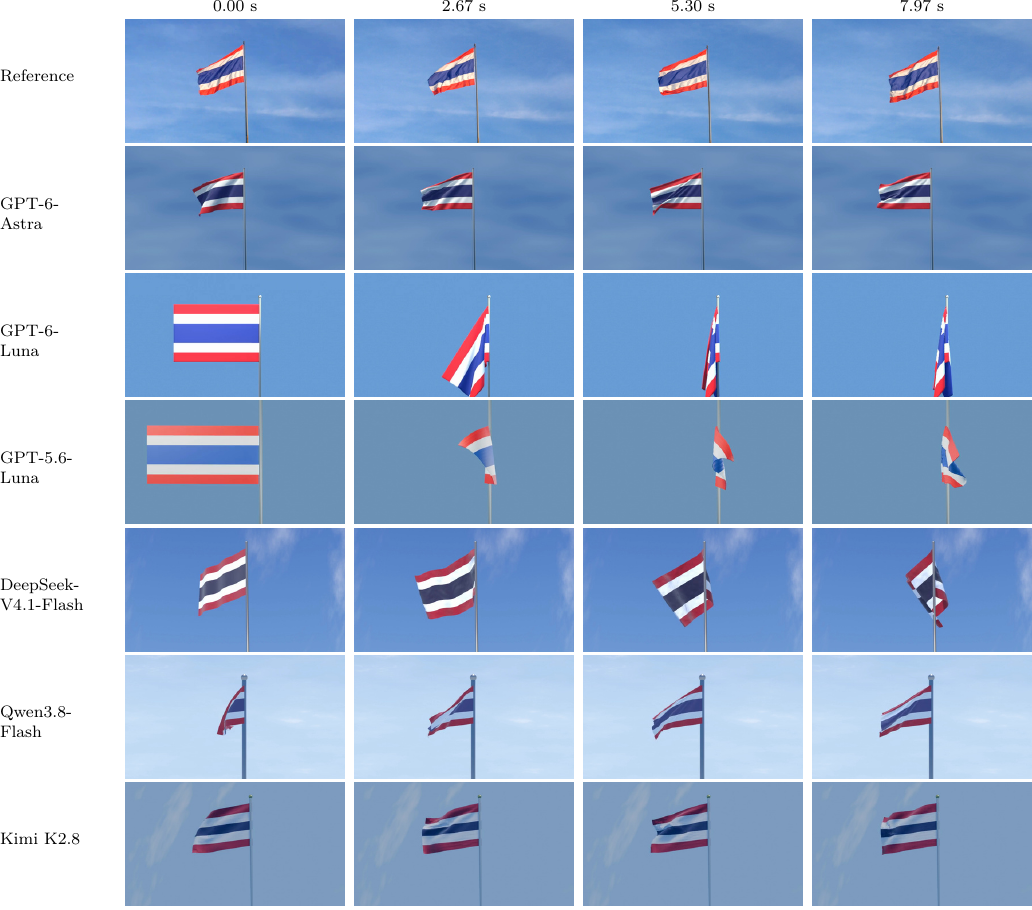}
\par\vspace{3pt}
\begingroup\fontsize{8}{9.5}\selectfont
\noindent\parbox{\linewidth}{\textbf{Component and summary scores.} Best and second-best scores are \textbf{bold} and \underline{underlined}, respectively.}\par\vspace{4pt}
\setlength{\tabcolsep}{0pt}
\renewcommand{\arraystretch}{1.18}
\begin{tabular*}{\linewidth}{@{}>{\raggedright\arraybackslash}p{76pt}@{\hspace{7pt}}>{\raggedright\arraybackslash}p{45pt}@{\hspace{7pt}}*{9}{>{\centering\arraybackslash}p{\dimexpr(\linewidth-135pt)/9\relax}}@{}}
\toprule
 & & \multicolumn{3}{c}{Visual metrics} & \multicolumn{3}{c}{Physical metrics} & \multicolumn{3}{c}{Summary scores} \\
\cmidrule(lr){3-5}\cmidrule(lr){6-8}\cmidrule(l){9-11}
{\fontsize{9}{10.5}\selectfont Model} & {\fontsize{9}{10.5}\selectfont Harness} & $S_{\rm SSIM}\!\uparrow$ & $S_{\rm LPIPS}\!\uparrow$ & $S_{\rm obj}\!\uparrow$ & $S_{\rm evt}\!\uparrow$ & $S_{\rm mot}\!\uparrow$ & $S_{\rm val}\!\uparrow$ & $S_{\rm vis}\!\uparrow$ & $S_{\rm phys}\!\uparrow$ & $S_{\rm tot}\!\uparrow$ \\
\midrule
GPT-6-Astra & Codex & \textbf{0.863} & \underline{0.711} & \textbf{0.833} & \textbf{1.000} & 0.231 & \textbf{1.000} & \underline{0.802} & \textbf{0.744} & \textbf{0.773} \\
GPT-6-Luna & Codex & 0.853 & 0.634 & \underline{0.667} & \underline{0.000} & 0.226 & \underline{0.500} & 0.718 & 0.242 & 0.480 \\
GPT-5.6-Luna & Codex & \underline{0.861} & 0.633 & \underline{0.667} & \textbf{1.000} & 0.174 & \textbf{1.000} & 0.720 & \underline{0.725} & \underline{0.722} \\
DeepSeek-V4.1-Flash & Claude Code & 0.859 & \textbf{0.725} & \textbf{0.833} & \textbf{1.000} & \underline{0.272} & \underline{0.500} & \textbf{0.806} & 0.591 & 0.698 \\
Qwen3.8-Flash & Claude Code & 0.801 & 0.683 & \textbf{0.833} & \textbf{1.000} & \textbf{0.395} & \underline{0.500} & 0.772 & 0.632 & 0.702 \\
Kimi K2.8 & Kimi Code & 0.857 & 0.695 & \textbf{0.833} & \textbf{1.000} & 0.191 & \underline{0.500} & 0.795 & 0.564 & 0.679 \\
\bottomrule
\end{tabular*}
\endgroup

\caption{\textbf{Event completion with cloth self-intersections.} Qwen3.8-Flash receives an event score of $1$, yet physical-validity checks detect self-intersections between nonadjacent cloth regions in $14$ frames, capping validity at $0.5$. This case shows that reproducing the visible event can coexist with invalid cloth geometry. Displayed frames also show differences in flag extension and cloth phase.}
\label{fig:diverse-thai-blue}
\end{figure}

\clearpage
\subsection{Physical Validity Failures}
\label[appendix]{sec:validity-showcase}

The six selected reconstructions receive $S_{\mathrm{obj}}=S_{\mathrm{evt}}=1$, yet have explicit physical-validity failures.
All displayed validity scores are resolved raw proposals from the original-condition checks; no default-pass assignments are used.
This subsection illustrates rubric criteria evaluated on unchanged scenes; counterfactual criteria appear in \cref{sec:counterfactual-showcase}.
Image pairs use the same reference/submission frame index. Plots identify one failure witness; scores summarize the corresponding original-condition rubric.
Pixel fidelity and timing vary; these examples do not estimate failure prevalence.

\begin{figure}[!ht]
\centering
\includegraphics[width=\textwidth]{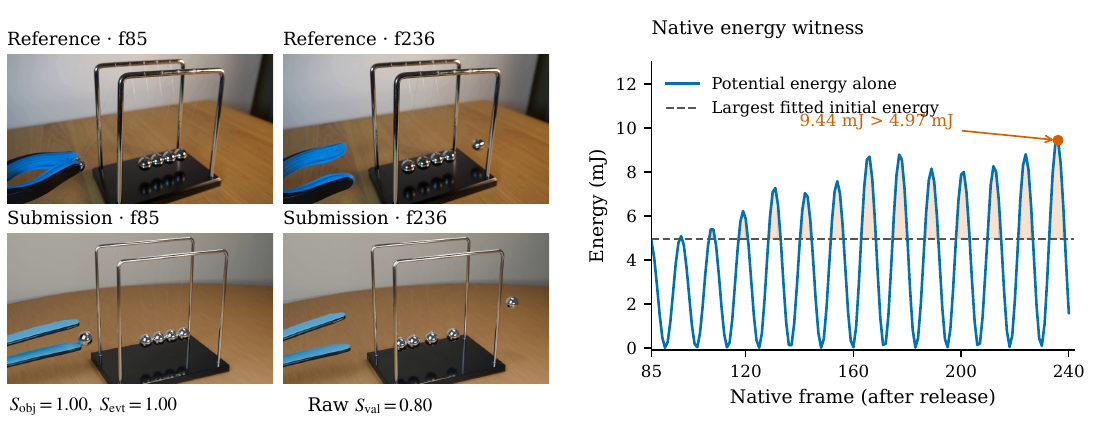}
\caption{\textbf{GPT-6-Astra: energy addition in a Newton cradle.}
After the gripper is clear at frame~85, four fits to the free pre-impact swing estimate initial total energy at approximately $4.971$\,mJ; the dashed line is the largest fitted estimate.
At frame~236, gravitational potential alone reaches $9.441$\,mJ.
With fixed supports and no admitted continuing work source, the native trajectory violates the passive energy budget (N3).
This witness does not require differentiating impact-frame velocities.}
\label{fig:validity-astra-energy}
\end{figure}

\begin{figure}[!ht]
\centering
\includegraphics[width=\textwidth]{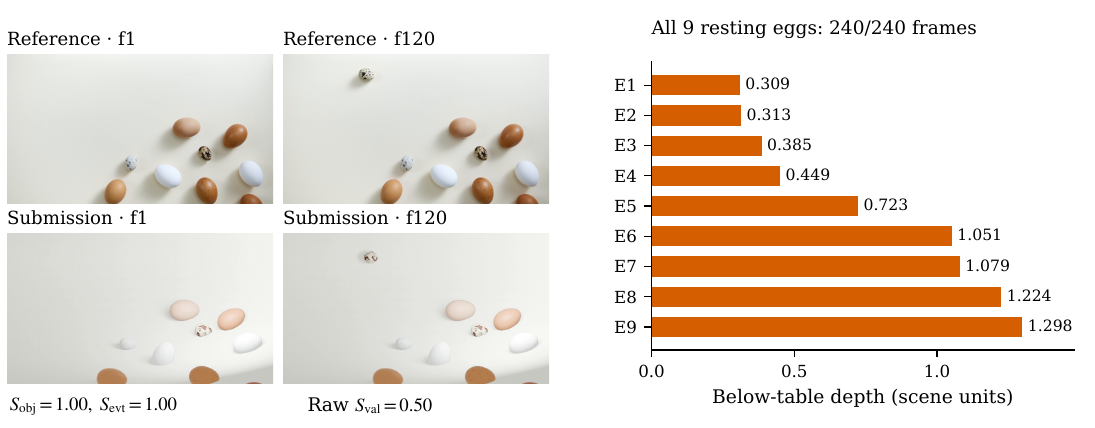}
\caption{\textbf{GPT-5.6-Sol: resting eggs embedded in the tabletop.}
All nine resting eggs intersect the actual evaluated table surface throughout $240/240$ frames (E1).
Their deepest below-surface vertices are $0.309$--$1.298$ scene units beneath the local surface, with persistent material-boundary intersections.
The bars describe the resting eggs only, excluding the mover's isolated initial overlap.}
\label{fig:validity-sol-eggs}
\end{figure}

\clearpage
\begin{figure}[!ht]
\centering
\includegraphics[width=\textwidth]{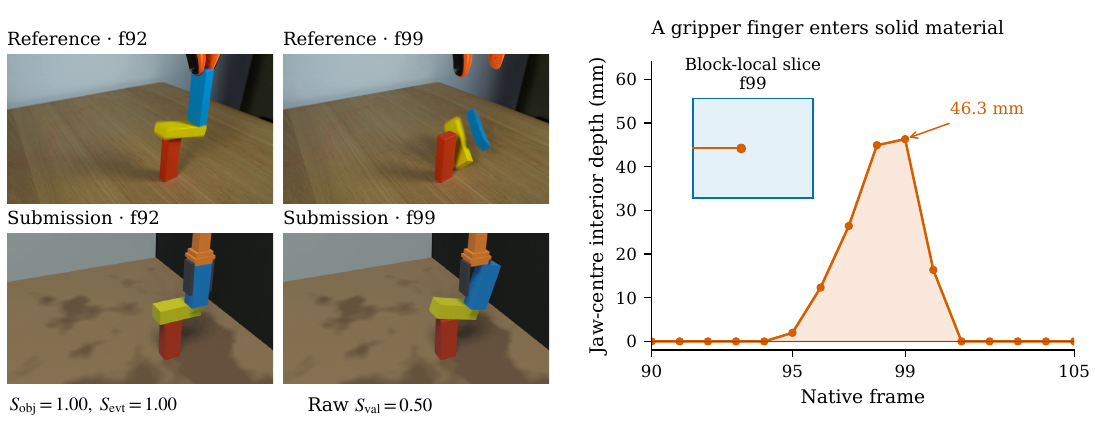}
\caption{\textbf{DeepSeek-V4.1-Flash: a gripper passes through a released block.}
The right jaw's centre lies inside the blue block on frames~95--100, reaching $46.3$\,mm from the nearest block face at frame~99 (K1).
The inset is a block-local cross-section through that centre, derived from measured coordinates.
The plotted quantity is centre interior depth, not a universal surface-penetration depth; the original-condition score also includes an independent guide-attachment failure.}
\label{fig:validity-deepseek-gripper}
\end{figure}

\begin{figure}[!ht]
\centering
\includegraphics[width=\textwidth]{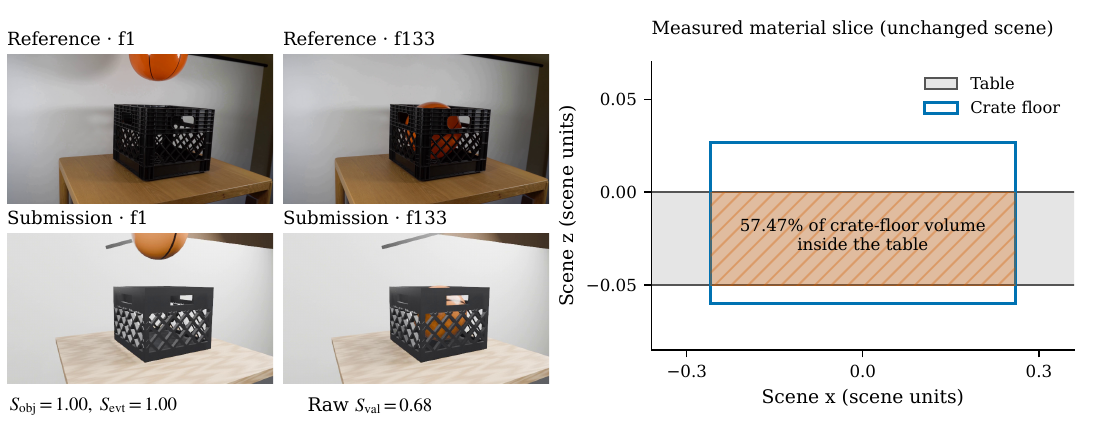}
\caption{\textbf{GLM-5.3-Flash: the crate floor occupies the wooden tabletop.}
The measured solid intervals overlap by $0.51864\times0.51864\times0.05000$ scene units, embedding $57.47\%$ of the crate-floor volume in the table (B1).
The right panel is a section through the actual axis-aligned box materials, with their common material shaded.
Both bodies remain unchanged throughout all $240$ frames; this is a persistent original-state overlap, not a failure inferred by modifying the scene.}
\label{fig:validity-glm-basket}
\end{figure}

\clearpage
\begin{figure}[!ht]
\centering
\includegraphics[width=\textwidth]{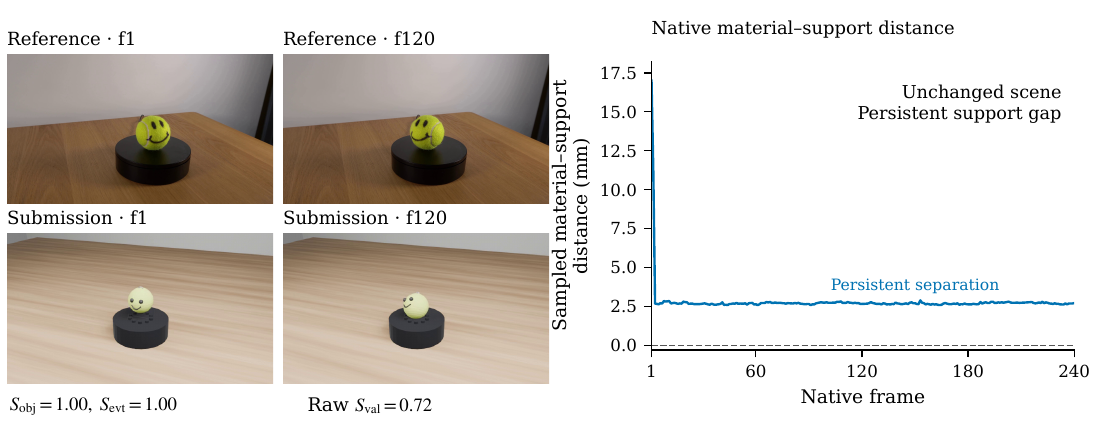}
\caption{\textbf{Claude Sonnet 5: rotation supported by an oversized collision proxy.}
Joined top decoration enlarges the spherical collider beyond the ball material, producing persistent remote support/traction (S2).
The main curve is the minimum over sampled vertices, edge midpoints and face centroids, not an exact global surface-distance minimum.
The unchanged scene continues to support and rotate the ball while sampled material distances remain positive; the dashed line denotes zero material separation.
Only original-state measurements are displayed.}
\label{fig:validity-sonnet-gap}
\end{figure}

\begin{figure}[!ht]
\centering
\includegraphics[width=\textwidth]{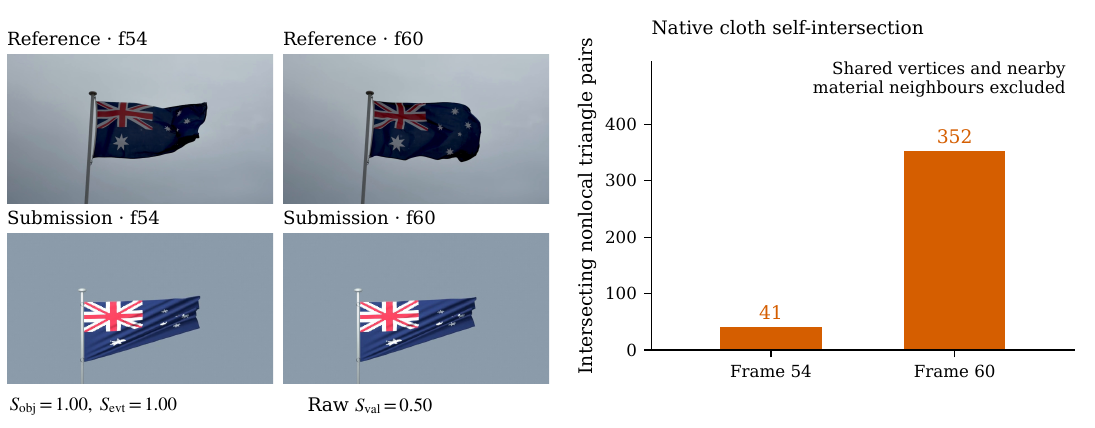}
\caption{\textbf{GPT-6-Luna: nonadjacent flag material crosses itself.}
Exhaustive checks of unchanged Blender-scene frames~54 and~60 find $41$ and $352$ intersecting nonlocal triangle pairs (F2).
Pairs sharing a material vertex or lying within a two-quad material neighbourhood are excluded; a frame-54 witness joins patches separated by $18$ material rows.
These are geometric pair counts at the two inspected frames, not independent events or a claim about the first onset of crossing.
The raw validity score is the resolved value $0.50$.}
\label{fig:validity-luna-flag}
\end{figure}

\clearpage
\subsection{Counterfactual Mechanism Checks}
\label[appendix]{sec:counterfactual-showcase}

A reconstruction can pass checks along its original trajectory while omitting a physical interaction that the trajectory never exercises.
The five selected reconstructions below illustrate this gap: a rolling body does not encounter a nearby dish, the cue ball remains on the table, or a bearing keeps a rotor away from the surface below.
Changing the release or connection activates the missing interaction.
A failure is established by measured material intrusion or unsupported motion together with a responsive control, rather than by trajectory change alone.

Each selected reconstruction scores $1.0$ on its original-condition checks but fails a counterfactual criterion of its physical-validity rubric.
The paired reports use identical source and saved-world hashes.
The native scene views show the tested intervention states, paired with measurements of geometry and motion.
The examples illustrate failures of counterfactual rubric criteria; they do not estimate a failure rate.

\subsubsection{A Nearby Obstacle Has No Contact Response}

\begin{figure}[!ht]
\centering
\includegraphics[width=\textwidth]{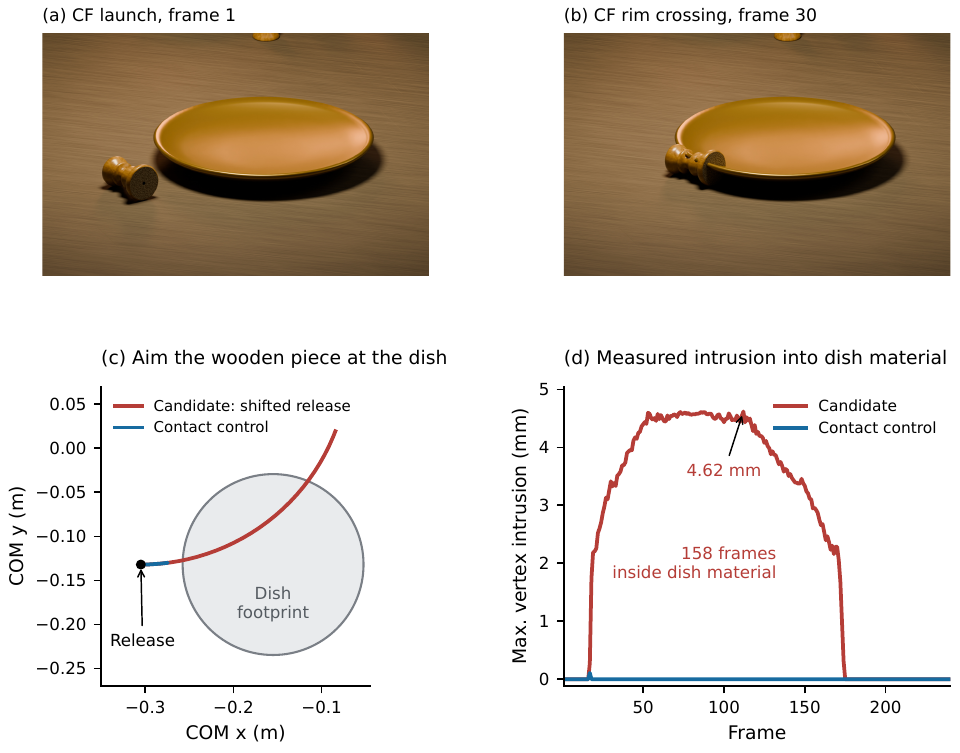}
\caption{\textbf{GPT-6-Astra: the wooden piece passes through the dish.}
The views show the shifted release before contact and during rim crossing.
The original release avoids the dish.
A shifted release starts $32.25$\,mm clear of its surface but puts piece vertices inside the actual closed dish material for $158$ frames, reaching $4.619$\,mm depth.
The grey disc in (c) is only the dish footprint; the failure witness in (d) measures intrusion into its three-dimensional material.
Removing the dish leaves the candidate's centre-of-mass trajectory unchanged.
An independent contact control using the source mesh, mass, inertia and integration scale arrests the approach, with only $0.110$\,mm maximum local contact intrusion.
Thus the issue is absent obstacle response, not merely overlap in the image or the top-down projection.}
\label{fig:cf-astra-wood}
\end{figure}

\clearpage
\subsubsection{Support Extends Beyond the Physical Table}

\begin{figure}[!ht]
\centering
\includegraphics[width=\textwidth]{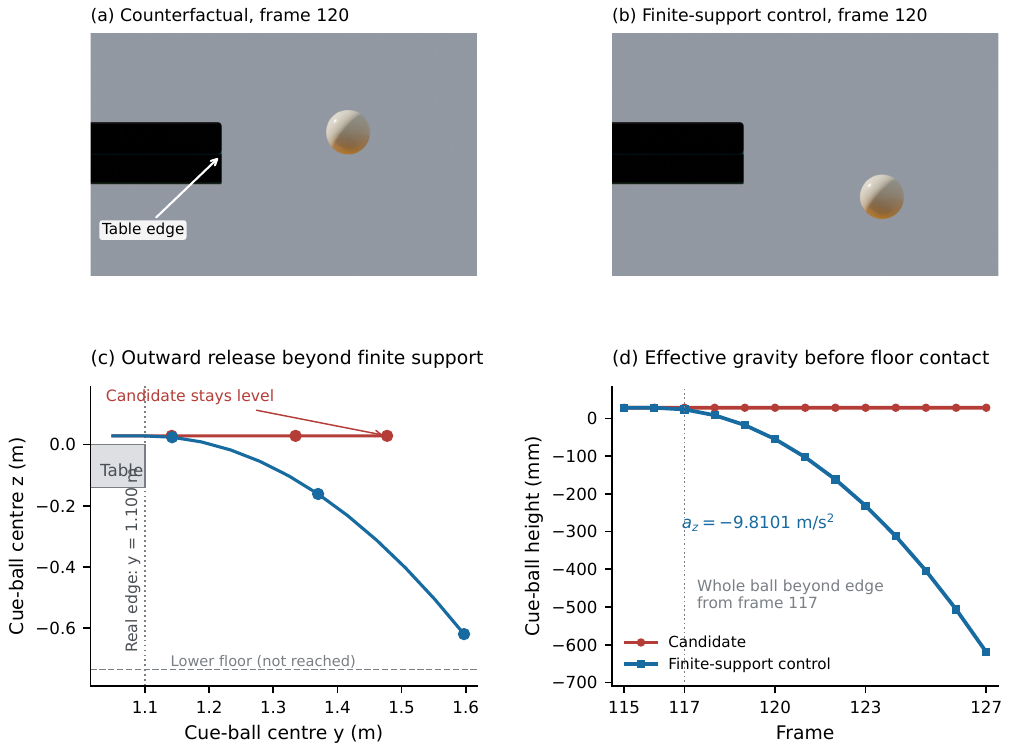}
\caption{\textbf{GPT-6-Astra: the cue ball remains suspended beyond the table edge.}
The views compare the outward-release candidate and finite-table control at frame~120.
An outward release moves the entire cue-ball envelope beyond the actual support boundary at $y=1.100$\,m from frame~117.
Its centre remains at approximately $28.575$\,mm height through frame~240 ($124$ samples), despite the absence of table material below it.
An explicit infinite-support control reproduces the candidate trajectory exactly.
Restricting support to the finite tabletop reproduces the prefix through frame~116 and then gives $a_z=-9.8101$\,m/s$^2$.
Only frames~115--127 are shown for that control, before it reaches the lower floor; its later floor-contact behavior is outside this witness.
The plotted table patch marks the top support boundary, not a scaled cross-section of the entire table.}
\label{fig:cf-astra-billiard}
\end{figure}

\paragraph{What the intervention distinguishes.}
The original-condition checks see rolling supported motion and do not exercise this upper boundary.
The intervention tests the spatial extent of the support law against the actual scene material.
The finite-support and infinite-support controls separate a missing boundary condition from an ineffective gravity change.
The defect is in the reconstructed support mechanism; it is not inferred from the ball leaving the camera view.

\clearpage
\subsubsection{A Bearing Hides a Missing Table Collision}

\begin{figure}[!ht]
\centering
\includegraphics[width=\textwidth]{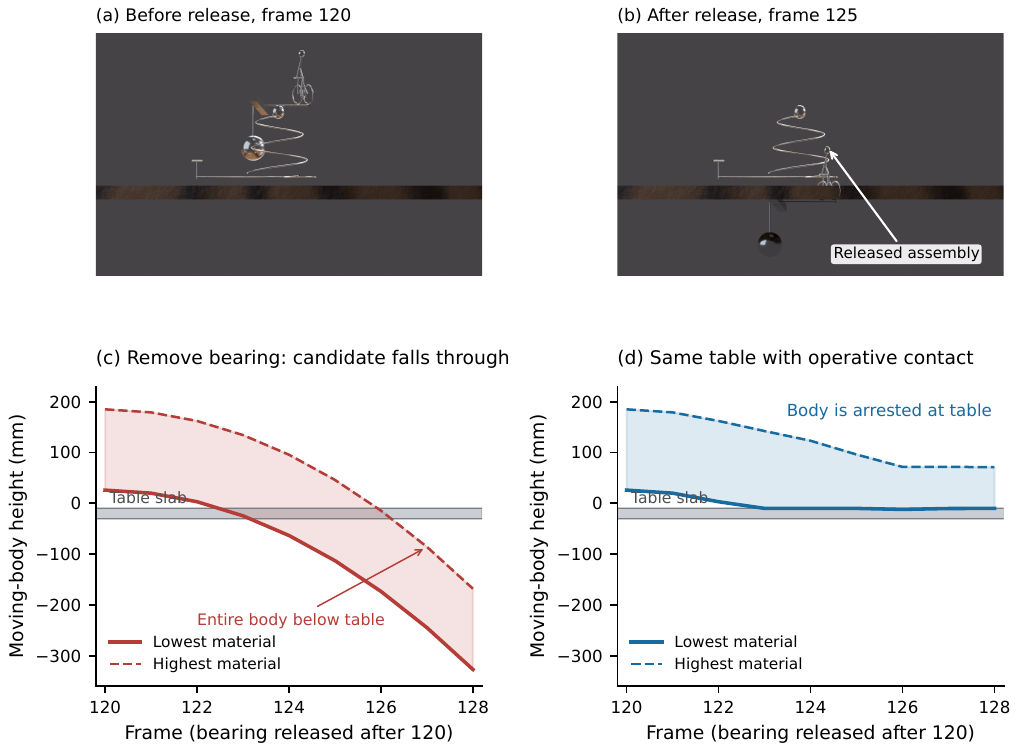}
\caption{\textbf{DeepSeek-V4.1-Flash (max effort): releasing the rotor exposes missing table contact.}
The views show the assembly just before release and intersecting the table at frame~125.
The original bearing supports the moving assembly.
The intervention retains the exact trajectory through frame~120, then removes the sole base-to-beam hinge.
The body intersects the unchanged table on frames~123--126 and lies entirely below the table bottom at frame~127.
The shaded red and blue bands show each branch's measured lowest-to-highest material extent, not a filled-body cross-section.
Giving the same table geometry an operative collision path arrests the body at its surface; table-removal and disabled-contact controls reproduce the candidate's free fall.
Falling after a bearing is removed is expected. Passing through the table is the counterfactual failure.}
\label{fig:cf-deepseek-wire}
\end{figure}

\paragraph{Why the control matters.}
Removing a legitimate support does not itself establish that a reconstruction is invalid.
Here, the bearing-retained branch and the unchanged prefix isolate the release, while the table-contact control demonstrates that the available surface can stop the body.
The intervention reveals a contact mechanism that the supported original motion did not require.
Full source identities, intervention records and trajectory data are retained with the figures.

\clearpage
\subsubsection{A Different Reconstruction Omits Contact with a Chess Piece}

\begin{figure}[!ht]
\centering
\includegraphics[width=\textwidth]{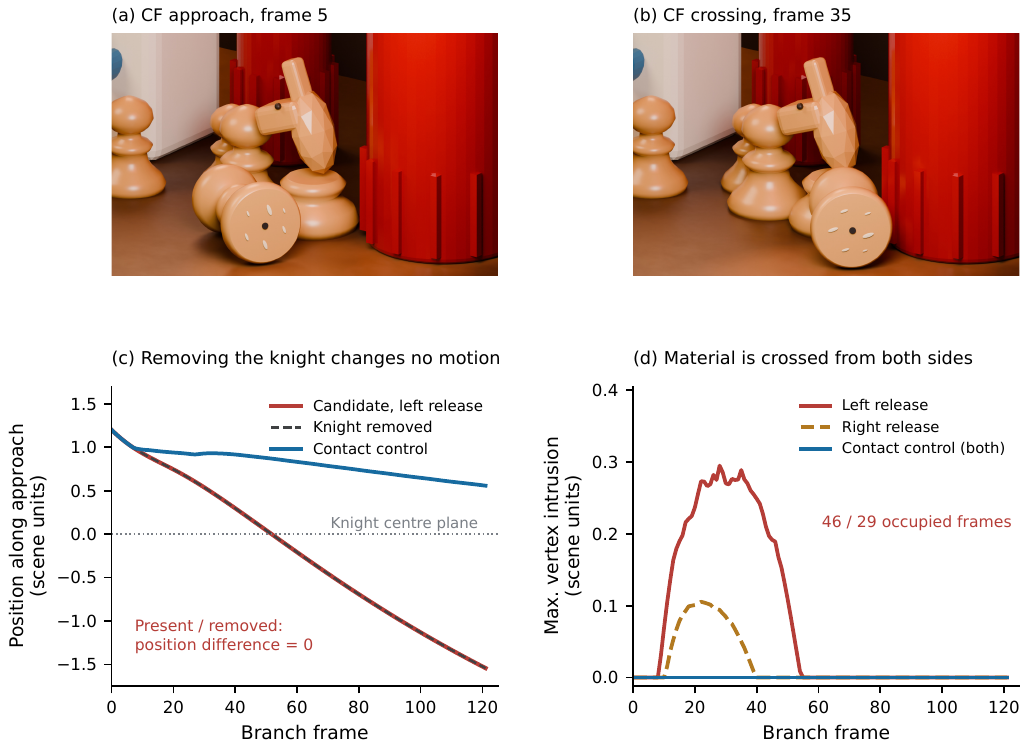}
\caption{\textbf{GPT-5.6-Sol: the rolling piece passes through the knight.}
The views follow the left-release branch before and during the crossing.
This reconstruction of the same rolling task contains a nearby closed knight assembly whose five material meshes have no collision owner.
Two initially separated, table-supported releases put the moving piece inside knight material for $46$ and $29$ frames, with maximum interior vertex depths of $0.295$ and $0.105$ scene units.
Depth is the largest measured vertex intrusion in either direction between the two meshes.
In (c), the left-release positions and velocities are exactly unchanged when the knight is removed, across all $122$ recorded samples; crossing the centre plane is only a path illustration, while (d) measures actual material intrusion.
Adding passive contact to the unchanged target geometry produces a response and avoids material intrusion for both releases.
Repeated runs and doubled solver substeps retain the failure.
Length values use native scene units.}
\label{fig:cf-sol-knight}
\end{figure}

\paragraph{What the original trajectory leaves untested.}
The original-condition checks certify the rolling body's rigidity, continuity and table support.
They do not require a response from the nearby knight because the original motion never reaches it.
The changed releases activate this missing contact path.
This extends the dish example in \cref{fig:cf-astra-wood} to a different submitted program and a different obstacle selected from its own scene.

\clearpage
\subsubsection{A Solid Token Behaves as if It Were Absent}

\begin{figure}[!ht]
\centering
\includegraphics[width=\textwidth]{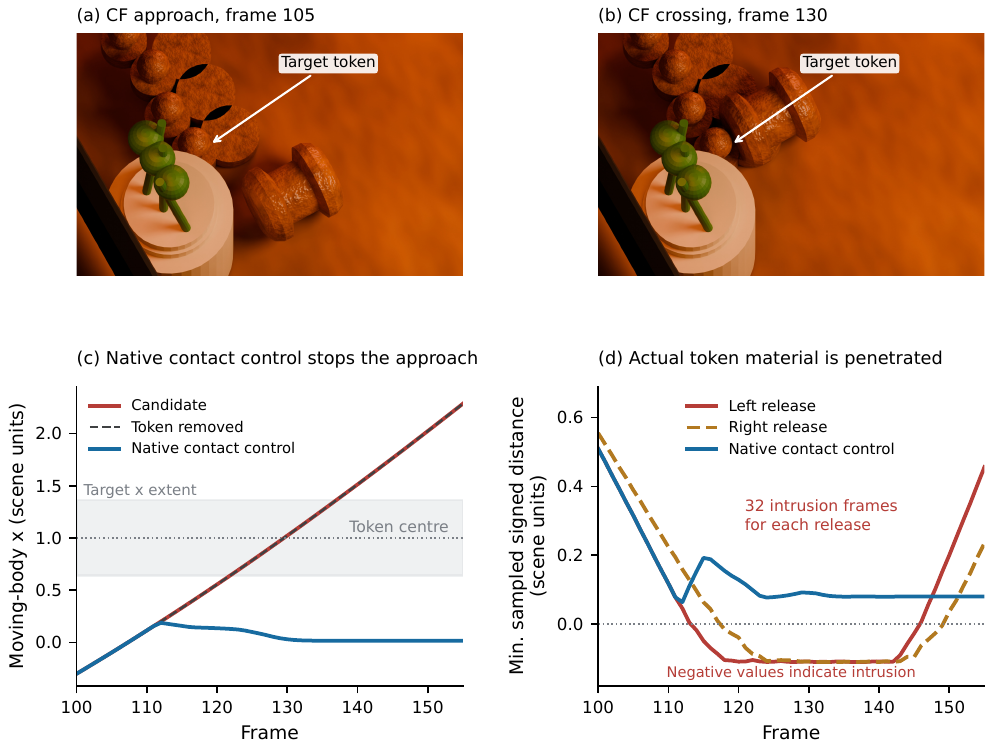}
\caption{\textbf{GPT-5.6-Luna: the rolling piece traverses a solid wooden token.}
The views track the left-release branch as the body enters the selected token.
The selected target is a filled cylinder with no hole and no physical contact owner.
Two target-reaching releases produce mesh intersections on frames~114--145 and~118--149, respectively: $32$ consecutive frames each.
The minimum cylinder signed distance over sampled moving-body surface points is approximately $-0.110$ scene units in both branches.
For the left release, removing the token changes none of the $240$ sampled positions.
A native contact control matches the prefix through frame~111 and then stops the approach without token penetration.
Its collider responds before the actual material surface, with a dense measured separation of $0.0622$ scene units; it is a sensitivity control, not an exact surface-contact trajectory.
The grey band in (c) is only the target's projected $x$ extent; (d) gives the material-distance witness.}
\label{fig:cf-luna56-token}
\end{figure}

\paragraph{Scope of the comparison.}
The original motion avoids this obstacle, so its table-support checks can pass while the obstacle remains physically inactive.
The intervention exposes the mismatch between rendered solid material and the program's operative collision scene.
The initial opposite-side placement blocked by another real object remains in the archived test matrix; the plotted opposite-side branch uses a separately measured, target-reaching release.
All three reconstructions of this rolling task retain their own geometry and source scale; no common artificial obstacle is inserted to create the failure.

\clearpage
\subsection{Cross-Harness Verification}
\label[appendix]{sec:cross-harness-cases}

Harness instructions may influence when an agent checks its reconstruction and how it reserves execution budget for corrections.
We explore this possible connection through two DeepSeek-V4.1-Flash tasks where Codex completes execution checks after its final revisions and obtains better event scores than Claude Code.
Both agents perform earlier tests; the difference is whether the final repairs are followed by another successful run and output inspection.
The galleries show videos generated by the benchmark verifier from the submitted programs, while \cref{sec:harness-trajectories} examines the agents' own verification steps and the harness guidance behind them.

\subsubsection{Checking Rolling After Contact Changes}

In the wooden-piece task, Claude Code performs prototype measurements but exhausts its Blender allowance before executing the final rewritten program.
Its last checks find no saved world or completed frame set, and the finalizer refuses the submission (steps~199--205).
Codex also initially produces a stalled object.
It detects the stall, tests collision shapes and table resolution, and revises the contact geometry and incline (132--160).
Its twentieth launch renders the final program, after which it inspects output frames and diagnostics and finalizes (163--169).

\begin{figure}[!ht]
\centering
\includegraphics[width=\textwidth]{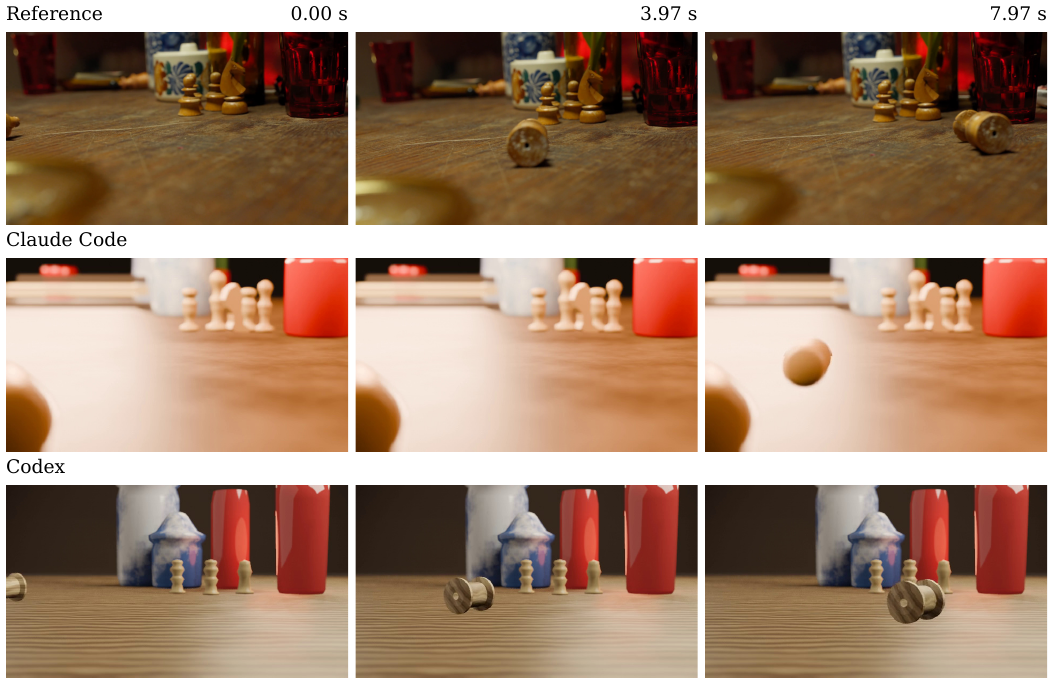}
\caption{\textbf{DeepSeek-V4.1-Flash: wooden-piece rolling.}
Codex tests and revises the contact setup before completing the final render and checking its motion.}
\label{fig:harness-wood}
\end{figure}

\paragraph{Outcome.}
The benchmark event score rises from $0$ for Claude Code to $1$ for Codex.
Appearance remains imperfect: both receive $S_{\mathrm{obj}}=0.286$.
Codex's tests are followed by contact changes, a new full run, and recovery of the rolling event.
Claude Code's final revision has static checks and earlier-prototype evidence, but no successful execution after the final changes.
This pattern suggests that completing behavioral checks within the launch budget may matter for event fidelity.

\clearpage
\subsubsection{Checking Accumulation Across a Pour}

In the wine-pouring task, Claude Code's final production bake aborts.
It then changes the solver substeps, initial-liquid clearance, and resolution, but has no remaining launch budget to execute those fixes; the agent's output directory remains empty and finalization fails (224--248).
Codex inspects sampled frames from a complete rendered sequence, identifies an excessive filling rate, and twice adjusts the inlet and reruns the program (80--90).
It then inspects separated frames and checks a saved-scene preview at frame~240 against the delivered frame (91--94).

\begin{figure}[!ht]
\centering
\includegraphics[width=\textwidth]{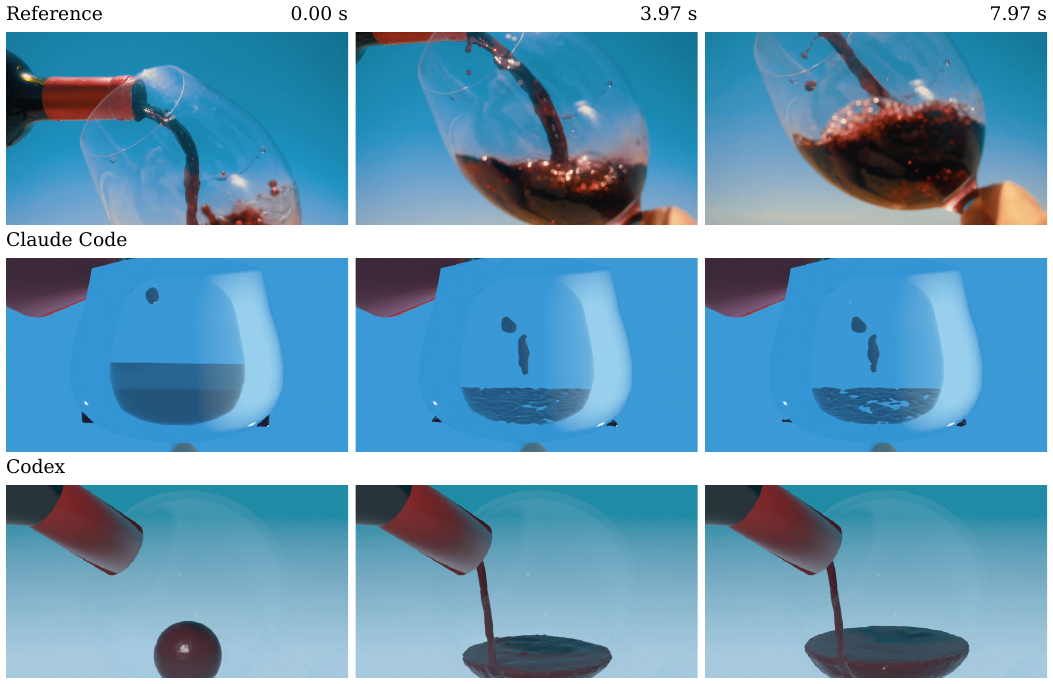}
\caption{\textbf{DeepSeek-V4.1-Flash: wine pouring.}
Codex's checks lead to repeated inlet corrections and a new full render after each correction.}
\label{fig:harness-wine}
\end{figure}

\paragraph{Outcome.}
The event score increases from $0$ to $0.857$, and $S_{\mathrm{vis}}$ from $0.521$ to $0.601$.
Codex still fails the event item requiring the tilted, visibly supported glass; not every metric improves.
Its frame-240 preview has zero pixel difference from the corresponding delivered frame, which establishes that single-frame agreement only.
The sequence illustrates how checks across time can reveal a filling-rate error and guide further corrections when execution budget remains.

\clearpage
\subsubsection{Harness Instructions and Completed Verification}
\label[appendix]{sec:harness-trajectories}

\paragraph{What the prompts require.}
The archived Codex~0.150.1 sessions explicitly instruct the agent, in the recorded \texttt{never} approval mode, to ``proactively run tests, lint and do whatever you need to ensure you've completed the task.''
Its task prompt also starts with a reminder to reserve the last Blender launch for final validation.
After removing this prefix, the benchmark task text is byte-identical across harnesses for both selected tasks: it asks the agent to build a minimal runnable scene, verify it, refine it, and run \texttt{finalize\_submission}.
Thus Claude Code also receives an explicit task-level verification requirement.

Version-matched local captures of Claude Code~2.1.274 and~2.1.278 emphasize faithful reporting of failed tests and skipped steps; we did not observe the same broad proactive-testing instruction in those captures.
This part of the comparison uses local reproductions because the historical full Claude Code system prompts were not retained.

\begin{table}[!ht]
\centering
\small
\setlength{\tabcolsep}{5pt}
\renewcommand{\arraystretch}{1.16}
\caption{Prompt evidence for the selected DeepSeek-V4.1-Flash configurations.}
\label{tab:harness-prompt-evidence}
\begin{tabular}{p{.17\textwidth}p{.375\textwidth}p{.375\textwidth}}
\toprule
Evidence & Claude Code & Codex \\
\midrule
Benchmark task & Explicitly requires verification and finalization. & Same task body; an additional leading reminder repeats the final-validation budget requirement. \\
Native prompt & Historical full prompt unavailable. Version-matched local captures emphasize faithful reporting of tests and skipped steps. & Historical session metadata records an explicit proactive-testing instruction under \texttt{never}. \\
Interpretation & Earlier checks are present, but the final revisions in these two cases remain unexecuted. & The final revisions in these two cases are run and inspected before the attempt ends. \\
\bottomrule
\end{tabular}
\end{table}

\paragraph{The observed verification sequences.}
\Cref{tab:harness-verification-sequences} connects the prompt comparison to concrete checks, corrections, and subsequent runs.
The two Codex traces contain no observed bypass of the launch cap, backdating of the submitted source, or substitution of older frames for the final revision.
The screenshots and scores describe verifier-stage videos; the table describes what the agents themselves managed to execute and inspect.

\begin{table}[!ht]
\centering
\small
\setlength{\tabcolsep}{5pt}
\renewcommand{\arraystretch}{1.16}
\caption{Final-revision verification in the two selected tasks. Step numbers refer to each case's own agent trace.}
\label{tab:harness-verification-sequences}
\begin{tabular}{p{.17\textwidth}p{.375\textwidth}p{.375\textwidth}}
\toprule
Task & Claude Code & Codex \\
\midrule
Wooden piece & Revises the final program after the cap is reached; static checks pass, but execution is blocked and output checks fail (166--205). & Detects stalling, revises contact settings, executes the final program at launch~20, then checks frames and finalizes (132--169). \\
Wine pouring & After a bake abort, changes numerical settings without a successful subsequent execution; finalizer reports missing frames (224--248). & Detects excessive accumulation, adjusts the inlet twice, reruns after each edit, inspects final frames, and checks a saved-scene preview (80--94). \\
\bottomrule
\end{tabular}
\end{table}

\paragraph{Possible connection to harness guidance.}
The event-score differences in these examples may be related to the harness's emphasis on proactive verification, the timing of behavioral checks, and the allocation of the limited launch budget.
Codex's instructions explicitly encourage verification, and its successful revisions here are followed by simulation and output inspection.
The two cases illustrate how such guidance may help an agent turn detected defects into tested corrections and better event reconstruction.

\end{document}